%% file: acl_latex.tex
\documentclass[11pt]{article}

\usepackage[preprint]{acl}
\usepackage{times}
\usepackage{latexsym}

\usepackage[T1]{fontenc}

\usepackage[utf8]{inputenc}

\usepackage{microtype}

\usepackage{inconsolata}

\usepackage{graphicx}
\usepackage{utfsym}
\usepackage{array}        
\usepackage{amsmath}
\usepackage{amssymb}
\usepackage{booktabs}   
\usepackage{pifont}     
\usepackage{makecell}   
\usepackage{multirow}     
\usepackage{multicol}     
\usepackage{colortbl}     
\usepackage[table]{xcolor} 
\usepackage{hhline}       
\usepackage{hyperref}
\usepackage{graphicx}     
\usepackage{tabularx}
\usepackage[ruled,vlined]{algorithm2e}
\usepackage{subcaption}
\newcommand{\cmark}{\textcolor{teal}{\ding{51}}} 
\newcommand{\xmark}{\textcolor{red}{\ding{55}}}
\definecolor{mygray}{gray}{0.92}
\title{
\raisebox{-0.8em}{\includegraphics[height=2.1em]{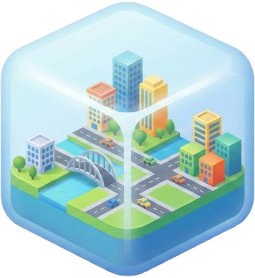}}\;
\textbf{CityCube: Benchmarking Cross-view Spatial Reasoning on Vision-Language Models in Urban Environments}
}

\author{
 \textbf{Haotian Xu\textsuperscript{1,2}},
 \textbf{Yue Hu\textsuperscript{1,2}},
 \textbf{Zhengqiu Zhu\textsuperscript{1,2}},
 \textbf{Chen Gao\textsuperscript{3}},
 \textbf{Ziyou Wang\textsuperscript{4}},\\
 \textbf{Junreng Rao\textsuperscript{1,2}},
 \textbf{Wenhao Lu\textsuperscript{1,2}},
 \textbf{Weishi Li\textsuperscript{1,2}},
 \textbf{Quanjun Yin\textsuperscript{1,2}},
 \textbf{Yong Li\textsuperscript{4}},
\\
 \textsuperscript{1}College of Systems Engineering, National University of Defense Technology,\\
 \textsuperscript{2}State Key Laboratory of Digital Intelligent Modeling and Simulation,\\
 \textsuperscript{3}BNRist, Tsinghua University,\\
 \textsuperscript{4}Department of Electronic Engineering, Tsinghua University
}

\begin{document}
\maketitle

\begin{figure*}[t]
\centering
\includegraphics[width=\linewidth]{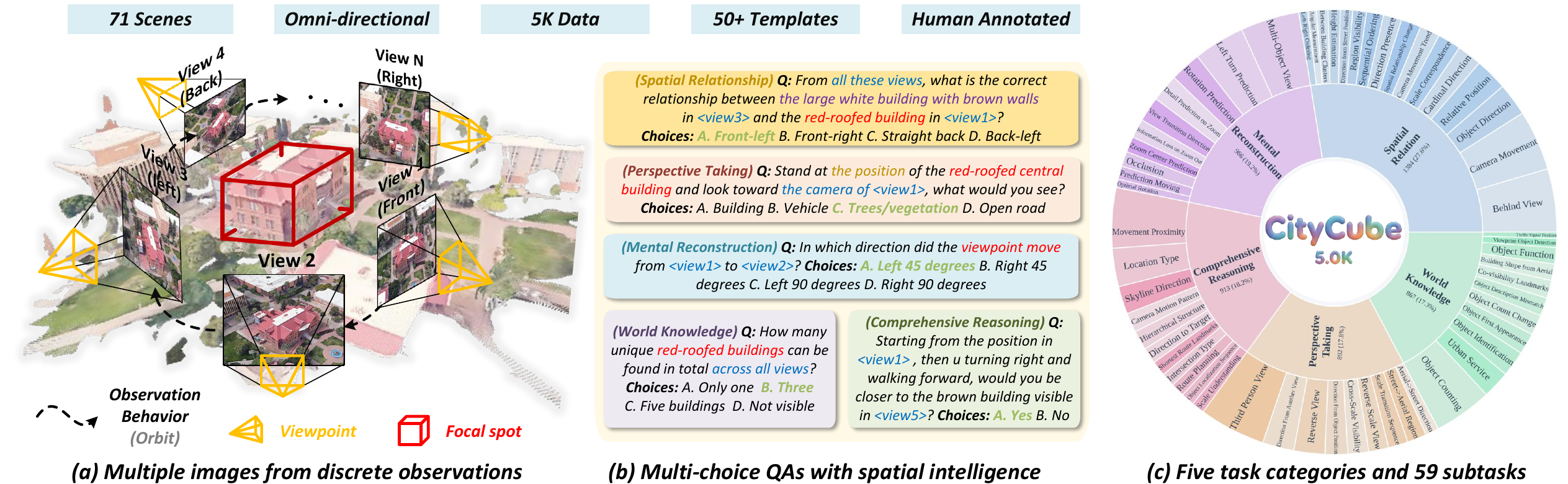}
\caption{Illustration of the CityCube benchmark. \textbf{Left:} An illustration of an embodied orbiting observation, where an agent captures multi-view images by circling a focal object (highlighted in red). \textbf{Middle:} Examples of multi-choice QA designed to evaluate five dimensions of CvSI. \textbf{Right:} Task distributions on CityCube Benchmark.}
\label{fig:teaser}
\end{figure*}

\input{sec/0_abstract}    
\input{sec/1_intro}
\input{sec/2_related_work}
\input{sec/3_method}
\input{sec/4_Experiments}
\input{sec/5_Conclusion}



\bibliography{custom}

\appendix

\input{sec/X_suppl}

\end{document}

%% file: sec/0_abstract.tex
\begin{abstract}
Cross-view spatial reasoning is essential for embodied AI, underpinning spatial understanding, mental simulation and planning in complex environments. 
Existing benchmarks primarily emphasize indoor or street settings, overlooking the unique challenges of open-ended urban spaces characterized by rich semantics, complex geometries, and view variations.
To address this, we introduce \textbf{CityCube}, a systematic benchmark designed to probe cross-view reasoning capabilities of current VLMs in urban settings. CityCube integrates four viewpoint dynamics to mimic camera movements and spans a wide spectrum of perspectives from multiple platforms, e.g., vehicles, drones and satellites. For a comprehensive assessment, it features 5,022 meticulously annotated multi-view QA pairs categorized into five cognitive dimensions
and three spatial relation expressions.
A comprehensive evaluation of 33 VLMs reveals a significant performance disparity with humans: even large-scale models struggle to exceed 54.1\% accuracy, remaining 34.2\% below human performance. By contrast, small-scale fine-tuned VLMs achieve over 60.0\% accuracy, highlighting the necessity of our benchmark.
Further analyses indicate the task correlations and fundamental cognitive disparity between VLMs and human-like reasoning.


\end{abstract}

%% file: sec/1_intro.tex
\section{Introduction}
\label{sec:intro}

Spatial reasoning across viewpoints and scales is fundamental to spatial intelligence. It goes beyond geometric measures \cite{yang2024depth, cai2025spatialbot}, also involves abilities such as relational reasoning \cite{chen2024spatialvlm, song2025robospatial}, perspective taking \cite{piaget2013child, li2025viewspatial}, mental simulation \cite{eslami2018neural}, dynamic perception \cite{tversky2019mind, ding2025understanding} and world knowledge recalling~\cite{jia2025omnispatial}. While humans naturally perform tasks like reasoning 3D scenes from streaming 2D views, replicating this in embodied AI remains challenging.


Recently, Vision-language Models (VLMs) have been increasingly adopted as the cognitive backbone for embodied agents, such as drones, mobile robots and autonomous vehicles~\cite{majumdar2024openeqa}. These agents dynamically operate and interact with the physical world, naturally requiring VLMs to perceive and understand multiple views~\cite{hong20233d, zhang2024agent3d, zhu2024llava, qi2024shapellm}. However, whether current VLMs possess the requisite capabilities for expansive urban spaces remains an open question, as comprehensive evaluations in this domain are still lacking.

To address this, we argue that benchmarks must extend beyond existing indoor-focused settings~\cite{yang2025thinking, du2024embspatial} to the broader context of urban open spaces. As shown in Fig.~\ref{fig:teaser}(a), urban environments present unique challenges for cross-view spatial intelligence (\textbf{CvSI}):



\begin{itemize}
    \item \textbf{Richer semantics:} Dense and repetitive instances (e.g., signage and vehicles) pose strict demands on VLMs in spatial grounding and disambiguation. 
    \item \textbf{Complex geometries:} Intricate urban 3D structures and road networks demand robust spatial mental reconstruction capabilities. 
    \item \textbf{Cross-scale viewpoint variations:} Diverse perspectives, from egocentric to top-down (e.g., drones), necessitate rigorous cross-scale reasoning to maintain spatial consistency.
\end{itemize}

To the best of our knowledge, CvSI in urban embodied tasks remains underexplored. To address this gap, we introduce \textbf{CityCube} (Fig.~\ref{fig:teaser}), a systematical benchmark for cross-view urban spatial reasoning.
CityCube is constructed by integrating urban views from real-world datasets and realistic simulators, forming a large-scale dataset with 18K images. It covers more than 70 representative cities, and includes multi-view imagery captured from heterogeneous platforms, providing a wide spectrum of first-person and aerial perspectives. 

Based on this foundation, we design a comprehensive and challenging suite of spatial reasoning tasks to systematically probe CvSI of urban embodiments.
Specifically, CityCube evaluates 5,022 QA pairs with five fundamental spatial intelligences in urban embodied scenes \cite{sensenova-si}, as illustrated in Fig.~\ref{fig:teaser}(b):  Spatial Relations (\textbf{SR}), Perspective Taking (\textbf{PT}), Mental Reconstruction (\textbf{MR}), World Knowledge (\textbf{WK}), and Comprehensive Reasoning (\textbf{CR}). Each ability requires cross-scale and multi-centric spatial reasoning over diverse scenes, reflecting the strengths and weaknesses of an agent in geometry, semantics, viewpoint transformation, and contextual knowledge.

Upon these tasks, we conduct a large-scale evaluation of 33 mainstream VLMs. As shown in Fig.~\ref{fig:teaser}(c), it reveals substantial performance gaps between models, persistent discrepancies with human reasoning, and limitations of existing spatial benchmarks in urban cross-view reasoning.
Beyond evaluation, CityCube is further split into training and testing sets. We fine-tune Qwen3-VL of variant scales on the training set using parameter-efficient LoRA, resulting in CityBot-2B, 4B and 8B. Experimental results show the potential of fine-tuning on CityCube in enhancing spatial reasoning, even for relatively small model scales.

In summary, our contributions are threefold:
\begin{itemize}
    \item \textbf{Dataset:} We introduce CityCube, a comprehensive dataset dedicated to cross-view spatial reasoning in urban embodied environments, covering 18.1K images from diverse perspectives across a wide spectrum of urban scenes. And this work further rearranges the imagery through four observation behavior primitives to mimic camera movements.
    \item \textbf{Benchmark:} We build a challenging CvSI benchmark including 5.0K QA pairs across 59 tasks under five fundamental cognition categories. The queries cover three kinds of relation expressions. Based on these problems, this work conducts a comprehensive evaluation over a diverse set of VLMs.
    \item \textbf{Findings:} We uncover several key findings, not limited to: (i)  significance of the benchmark, on which leading proprietary and open-source VLMs exhibit lower than 54.1\% accuracy, substantiating the challenge of CvSI, (ii) and disparity between current VLMs and human-like reasoning from correlation analyses and case studies.
\end{itemize}

%% file: sec/2_related_work.tex
\begin{table*}[t]
\centering
\caption{Comparison of the proposed and popular benchmarks for multi-view spatial intelligence. ``/'' indicates information not mentioned or included, ``+'' represents sequential operation. ``Total Tasks'' represents the amount of the task type. \textit{\textbf{Abbreviations}-- \textbf{Ped.}: Pedestrian, \textbf{Veh.}: Vehicle, \textbf{Sat.}: Satellite, \textbf{Tem.}: Templates, \textbf{Ego.}: Egocentric, \textbf{Allo.}: Allocentric, \textbf{Exo.}: Exocentric.} }
\label{tab:benchmark_comparison}

\resizebox{\textwidth}{!}{%
\small
\begin{tabular}{lcccccccccc}
\toprule

\textbf{Benchmark} & 
\textbf{Platform} & 
\textbf{\makecell{Camera \\ Orientation}} & 
\textbf{\makecell{QA \\ Num.}} & 
\textbf{\makecell{Reasoning \\ Annotation}} & 
\textbf{Annotation} & 
\textbf{Environment} & 
\textbf{\makecell{Embodied \\ Questions}} & 
\textbf{\makecell{Cross \\ Scale}} & 
\textbf{\makecell{CvSI \\ Categories}} & 
\textbf{\makecell{Total \\ Tasks}} \\

\midrule

All-Angles \cite{yeh2025seeing} & Pedestrian & Ego & 2.1K & \xmark & Human & Indoor & \cmark & \xmark & SR, PT & 6 \\
MMSI Bench \cite{yang2025mmsi} & Pedestrian & Ego, Exo & 1K & \cmark & Human & Indoor & \cmark & \xmark & \makecell[t]{SR, MR, PT} & 50 \\
ViewSpatial Bench \cite{li2025viewspatial} & Pedestrian & Ego, Exo & 5.7K & \xmark & Rules+Tem. & Indoor \& Web images & \cmark & \xmark & \makecell[t]{SR, PT} & 5 \\
MindCube \cite{yin2025spatial} & Pedestrian & Ego, Exo & 21.1K & \cmark & Rules+Tem. & Indoor & \cmark & \xmark & MR, PT & 5 \\
Ego3D Bench \cite{gholami2025spatial} & Vehicle & Ego & 8.6K & \cmark & Rules+Tem. & Outdoor driving & \cmark & \xmark & WK & 10 \\
OmniSpatial \cite{jia2025omnispatial} & Ped./Vehicle & Ego & 1.5K & \cmark & Human & Web images & \xmark & \xmark & PT & 50 \\
UrbanFeel \cite{he2025urbanfeel}& Pedestrian & Ego & 14.3K & \xmark & VFM+Human & Street views & \xmark & \xmark & / & 11 \\
Urbench \cite{zhou2025urbench}& Ped./Satellite & Ego, Allo & 11.6K & \xmark & Rules+LLM+Human & Satellite+Street & \xmark & \cmark & / & 14 \\

\midrule

\rowcolor{mygray}
\textbf{CityCube} & 
\makecell[c]{\textbf{Multi-source} \\ \footnotesize{(Ped., Veh., Drone, Sat.)}} & 
\makecell[c]{\textbf{Multi-centric} \\ \footnotesize{(Ego, Exo, Allo)}} & 
\textbf{5.0K} & 
\cmark & 
Tem.+LLM+Human & 
\makecell[c]{Urban aerial \\ \& street views} & 
\cmark & 
\cmark & 
\makecell[c]{SR, PT, MR, \\ CR, WK} & 
\textbf{59} \\

\bottomrule
\end{tabular}%
}
\end{table*}

\section{Related Work}
\label{sec:formatting}

\textbf{VLM Spatial Intelligence Benchmark}
VLMs have demonstrated significant potential in depth estimation and spatial cognition on spatial intelligence benchmarks like VSI-Bench \cite{yang2025thinking, cai2025spatialbot, cheng2024spatialrgpt, gholami2025spatial}. However, existing benchmarks \cite{yin2025spatial} often overlook the inherent cross-view nature of urban environments, limiting their evaluation scope to a broader applications.
While prior works such as ViewSpatial \cite{li2025viewspatial}, UrbanVideo \cite{zhao2025urbanvideo} and Urbench \cite{zhou2025urbench} have explored embodied urban spaces to some extent, they primarily focus on single or restricted perspectives, such as bird's-eye views. As shown in Tab \ref{tab:benchmark_comparison}, \textit{CityCube provides a unique and comprehensive benchmark by integrating multi-source imagery with multi-centric perspective modeling, and supports the richest set of task types with well-annotated reasoning processes}.\\
\textbf{Urban Visual Question Answering}
Urban Visual Question Answering (VQA) serves as a critical bridge between multi-modal learning and urban informatics. Unlike traditional urban QA systems that rely on retrieving structured information from static databases \cite{feng2025citygpt, feng2024citybench}, urban VQA emphasizes active perception through multi-modal cues, including vision and language. Existing works focus on perception at different scales. As shown in Tab \ref{tab:benchmark_comparison}, tasks at the macro level encompass geo-spatial querying, urban governance~\cite{zhou2025urbench, feng2025urbanllava}, and socioeconomic analysis~\cite{he2025urbanfeel, liu2025citylens, hao2024urbanvlp}. At the micro level, the focus shifts to agentic perception, including object localization~\cite{zhang2025open3dvqa} and motion planning~\cite{zhao2025urbanvideo}. \textit{However, current benchmarks still exhibit limitations in evaluating multi-image understanding, particularly for tasks requiring advanced cognitive abilities like cross-view reconstruction and perspective transformation in embodied contexts}. 

%% file: sec/3_method.tex
\section{CityCube Benchmark}
\label{sec:bench}

\subsection{Overview}
\begin{figure*}[t]
\centering
\includegraphics[width=\linewidth]{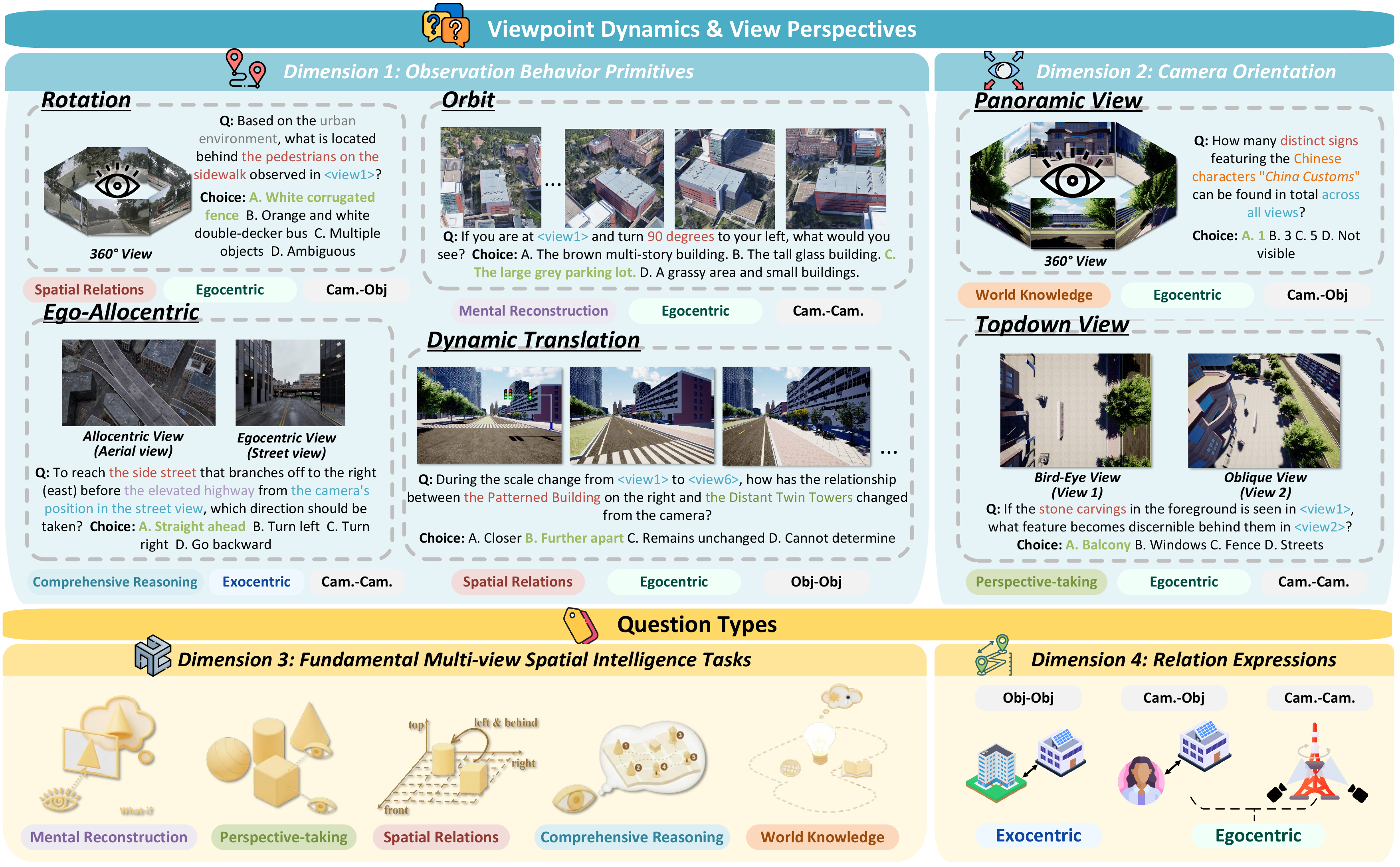}
\caption{The systematic evaluation protocol of the CityCube benchmark. \textbf{Upper Left:} Dim 1 evaluates observation with four representative behavior; \textbf{Upper Right:} Dim 2 tests model across various camera orientations; \textbf{Bottom Left:} Dim 3 categorizes 59 tasks into 5 fundamental categories; and \textbf{Bottom Right:}  Dim 4 labels the QA pairs with spatial reference frames.}
\label{fig:Overview}
\end{figure*}
As depicted in Fig.~\ref{fig:teaser}, CityCube targets multi-scale spatial reasoning under partial observability and dynamical urban viewpoints.
This benchmark is built on extensive real-world urban images (collected from 74 cities across the globe, including Singapore and Boston) and virtual images from 2 high-fidelity urban simulators (i.e. EmbodiedCity, and MatrixCity). In total, it offers \textbf{18.1K observation points} spanning diverse viewpoints, scales, and scene compositions. Building on this image pool, we carefully curate \textbf{5.0K QA pairs} to form the dataset, as shown in Fig.~\ref{fig:teaser}(c).

Specifically, the benchmark are structured into four well-designed dimensions, as shown in Fig.~\ref{fig:Overview}. The images are rearranged four different observation behavior primitives (\textbf{dim1}) with three camera perspectives (\textbf{dim2}). On this basis, we further identify five critical task categories (\textbf{dim3}) and define three spatial relations (\textbf{dim4}) for different queris to assess the VLMs from different CvSI abilities. 
Structurally, the first two determine the visual perspective and content while collecting the images, while the latter two dimensions are manifested in the textual QA pairs. Together, these variables configure the multi-image QA instances, effectively modeling the evaluation of high-level abilities as a superposition of targeted dimensions. The dataset statistics are illustrated in Appendix \ref{subsec:dataset_visualize}.

\subsection{Dimension 1: Viewpoint Dynamics}
To simulate active perception, we classify collected images to a set of behavior primitives: \textbf{(1) Rotation:} hovers and observes at a fixed position, changing only the camera orientation (yaw/pitch). \textbf{(2) Orbit:} surrounds a landmark or interested region, including planar and volumetric motions. \textbf{(3) Ego-Allocentric View:} simulates cognitive alignment from first-person views (egocentric) to third-person perspectives (allocentric). \textbf{(4) Dynaminc Translation:} mimics coarse-to-fine movements, such as zooming in from a skyscraper to a billboard. It involves multiple scales of urban space.

\subsection{Dimension 2: View Perspectives}
To ensure robustness across diverse embodiments, we standardize three acquisition protocols: \textbf{(1) Ground-level Panorama} represents views from ground vehicles (e.g., autonomous cars), covering discrete directions (front, rear, left, right, and diagonals) \cite{gholami2025spatial}. \textbf{(2) Low-altitude Oblique Imagery} represents views from aerial agents like drones, commonly used in 3D urban reconstruction to capture structural facades. \textbf{(3) Bird-Eye View} represents a global perspective with minimal occlusion, typical of satellite imagery or high-altitude mapping.

\subsection{Dimension 3: Benchmark Tasks}

CityCube establishes five fundamental CvSI task categories, built upon existing multi-view reasoning benchmarks in Tab \ref{tab:benchmark_comparison} while extending coverage to underexplored dimensions \cite{cai2025has}: 

\textbf{(1) Mental Reconstruction} requires VLMs to infer spatial transformation between views by mentally simulating hypothetical movement through the environment.

\textbf{(2) Perspective Taking} assesses spatial consistency maintenance of VLMs, focusing on cross-view object grounding and inferring relationship shifts as viewpoints change.

\textbf{(3) Spatial Relation} targets the precise quantification of estimating the distance, direction, and topological relations between objects observed across different viewpoints.

\textbf{(4) Comprehensive Reasoning} demands \textbf{multi-step spatial inference} for a hypothesis testing. For example, hypothetical navigation requires VLMs to mentally execute actions starting from a specific view and verify the predicted targets referencing other observed views.

\textbf{(5) World Knowledge} probes urban commonsense of VLMs, such as object geometry, affordance, and visibility. Beyond that, we also design challenges for recalling co-visible landmarks, object counting and scene captions across views.

\subsection{Dimension 4: Spatial Relation Expressions}
To unify spatial reference frames in QA expressions, CityCube tasks utilize three kind of binary spatial relations: \textbf{(1) Object-to-Object (Exocentric):} textual geometric relationships between external objects viewed from a spectator perspective.
\textbf{(2) Camera-to-Object (Egocentric):} spatial relations between the agent’s viewpoint and an observed target. \textbf{(3) Camera-to-Camera:} relative transformations between two distinct observation positions (e.g., Camera 1 vs. 2).

Rooted in these four critical dimensions, this design facilitates the construction of a systematic, interpretable, and fine-grained diagnostic benchmark. This benchmark is specifically engineered to model and evaluate high-level spatial reasoning challenges, including multi-view observation integration, geometric consistency under motion, spatial reference frame alignment, and object identity preservation across scale transitions.
\section{Benchmark Curation}
This section presents how this work constructs the image dataset and benchmark as depicted in Sec.~\ref{sec:bench}, including two main parts: image processing and QA generation.

\subsection{Data Collection and Pre-processing}
As shown in Fig.~\ref{fig:pipeline}, we collect urban images that satisfy predefined patterns and explicitly organize them into multi-view sets. The detailed image source are given in the Appendix \ref{subsec:images}. In pre-processing, we adopt two distinct pipelines for real-world datasets and 3D simulators, respectively. For real world, we design a multi-step image filtering and matching procedure to construct high-quality multi-view observations. For simulators, we manually collect viewpoint samples, record camera poses and motion trajectories. Across both real-world and simulated data, we ensure that multiple views correspond to the same underlying scene.

\textbf{Real-world scenes.}
All real-world images are sourced from public academic datasets. To obtain ground-level panoramic views mentioned in dim 2, we collect images from nuScenes \cite{caesar2020nuscenes}, which covers urban environments in Singapore and Boston. Since driving recording contains a large number of duplicate frames, we apply temporal frame skipping and image-similarity deduplication. In addition, GeoText-1652 \cite{chu2024towards} provides satellite and drone imagery captured from arbitrary viewpoints across 72 cities worldwide. For this dataset, we manually design sampling intervals to obtain approximate orbital observations in dim 1 and further remove redundant images based on image similarity. Detailed data sources and sampling strategies are provided in Appendix \ref{subsec:images process}.

\textbf{Simulated scenes.}
To further expand the scale and diversity of the dataset, we collect additional urban imagery from open-source simulators. EmbodiedCity \cite{gao2024embodiedcity} is a 3D urban simulator modeled after Beijing, containing multi-scale city elements ranging from large landmarks and commercial buildings to fine-grained objects such as bicycles and billboards. Based on this environment, we manually record agent trajectories over large areas following two representative motion patterns mentioned in dim 1 to support dynamic translation views, such as ``taking off'' (vertical movement) and ``approaching coffee shops'' (horizontal movement), which are critical for cross-scale spatial reasoning. Besides, we also supplement the dataset with a set of panoramic views captured by piloting the drone. 
MatrixCity \cite{li2023matrixcity} is a large-scale aerial-street view dataset from a virtual city. We compute geometric projections using precise camera poses to ensure view consistency between aerial and ground-level images, and manually filter image pairs that satisfy ego-allocentric views in dim 1 and 2. More detailed descriptions are provided in Appendix \ref{subsec:manual_images}.

\subsection{Question-Answer Generation}
To enrich multi-choice QA with CvSI tasks, we feed the processed image sets together with structured context into Gemini-2.5 Pro. The generation relies on three key context engineering strategies:

\textbf{(1) Contextual Role-Playing:} we prompt the VLM to serve as an urban embodiment, along with background knowledge of observation behaviors and perspectives. The specifications of prompts are listed in Sec \ref{subsec:role_playing}.

\textbf{(2) Template Coverage:} we also provide 59 distinct templates for formatting CvSI tasks, ensuring comprehensive coverage across all structural dimensions of the benchmark.

\textbf{(3) Geometric Reference Injection:} we explicitly supply ground-truth geometric information including camera position, orientation angle and image amount into the context. It is vital for model-based QA generation process, enhancing answer credibility, and mitigating hallucinations that violate physical constraints.

    

To mitigate potential construction biases arising from model, we implement a rigorous two-stage refinement pipeline as follows:

\textbf{(1) Blind Filtering:} we improve blind filtering \cite{zhao2025urbanvideo} for a text-only validation to mitigate textual biases. Specifically, an ensemble of models are used to assess questions without visual input. Questions are scored and stratified into difficulty levels (easy, moderate, hard etc) based on accuracy. By eliminating trivial samples, we obtain a QA dataset with a balanced and hierarchical difficulty distribution.

\textbf{(2) Human Verification:} our annotators verify the rationality and authenticity of the QA pairs, accepting or rejecting entries accordingly. Accepted questions undergo further proofreading to resolve ambiguities, invalid options, or erroneous reasoning, with human-authored reasoning processes added to enrich the annotations. The details are depicted in Sec \ref{sec:humanrefine}.

%% file: sec/4_Experiments.tex
\section{Experiments}
\subsection{Evaluation Setups}
\textbf{Evaluated Models.} As shown in Tab \ref{tab:complete_comparison}, apart from our fine-tuned models, we also evaluate the performance of 25 models under the CityCube benchmark, including 6 state-of-the-art proprietary models, 16 mainstream open-source models, and 3 specialized VLMs trained for spatial reasoning. 
Notably, the three spatial models are trained on respective spatial intelligence benchmarks, i.e., Spatialvlm \cite{chen2024spatialvlm}, Omni-spatial \cite{jia2025omnispatial}, and SSRL \cite{liu2025spatial}.

\textbf{Evaluation Protocol.} Leveraging the multiple-choice format, we compute task-level and overall average accuracy straightforwardly. For the human baseline, we recruited two independent groups of ten participants each, all with
urban science-related academic backgrounds (master’s or doctoral students). One
group conducted verifications in Sec 4.2, while the other performed the evaluations, ensuring no overlap between the two. More implementation details of the evaluation protocol are described in Appendix \ref{sec:humaneval}.

\subsection{Main Results}
\begin{table*}[t]
\scriptsize 
\centering
\caption{Accuracy of 33 VLMs on overall QA pairs. \textbf{Only three selected tasks are displayed for each CvSI category besides the overall accuracy due to space limitations.} The best performing model in each category is highlighted \textbf{in-bold}, while the second-best is \underline{underlined}.}
\label{tab:complete_comparison}
\renewcommand{\arraystretch}{1.15} 
\setlength{\tabcolsep}{1.5pt} 

\newcolumntype{Y}{>{\centering\arraybackslash}X}

\begin{tabularx}{\textwidth}{r|cc|YYYY|YYYY|YYYY|YYYY|YYYY}
    \hline
    & \multicolumn{1}{l}{} & \multicolumn{1}{l|}{} 
    & \multicolumn{4}{c|}{\cellcolor{cyan!20}\textbf{World Knowledge}} 
    & \multicolumn{4}{c|}{\cellcolor{green!20}\textbf{Perspective Taking}} 
    & \multicolumn{4}{c|}{\cellcolor{red!20}\textbf{Spatial Relation}} 
    & \multicolumn{4}{c|}{\cellcolor{orange!20}\textbf{Mental Recon.}} 
    & \multicolumn{4}{c}{\cellcolor{yellow!20}\textbf{Comp. Reasoning}} \\ 

    \textbf{Method} & \multicolumn{1}{l}{\textbf{Rank}} & \multicolumn{1}{l|}{\textbf{Avg.}} 
    & \rotatebox{90}{\textit{Overall Acc.}} 
    & \rotatebox{90}{\textit{Urban Service}} 
    & \rotatebox{90}{\textit{Object Ident.}} 
    & \rotatebox{90}{\textit{Object Counting}} 
    & \rotatebox{90}{\textit{Overall Acc.}} 
    & \rotatebox{90}{\textit{Another-view Dir.}} 
    & \rotatebox{90}{\textit{Third Person View}} 
    & \rotatebox{90}{\textit{Reverse View}} 
    & \rotatebox{90}{\textit{Overall Acc.}} 
    & \rotatebox{90}{\textit{Object Direction}} 
    & \rotatebox{90}{\textit{Relative Pos.}} 
    & \rotatebox{90}{\textit{Camera Move.}} 
    & \rotatebox{90}{\textit{Overall Acc.}} 
    & \rotatebox{90}{\textit{Multi-Obj View}} 
    & \rotatebox{90}{\textit{Rotation Pred.}} 
    & \rotatebox{90}{\textit{Left-turn Pred.}} 
    & \rotatebox{90}{\textit{Overall Acc.}} 
    & \rotatebox{90}{\textit{Route Planning}} 
    & \rotatebox{90}{\textit{Target Direction}} 
    & \rotatebox{90}{\textit{Location Type}} \\ 
    \hline

    
    \rowcolor[HTML]{ECF4FF} 
    \multicolumn{1}{l|}{\cellcolor[HTML]{ECF4FF}\textit{Baseline}} & & & & & & & & & & & & & & & & & & & & & & \\
    Random & - & 22.8 & 19.2 & 24.0 & 3.19 & 24.4 & 20.5 & 18.3 & 21.9 & 11.5 & 25.5 & 25.0 & 25.2 & 25.0 & 22.0 & 25.1 & 15.0 & 24.2 & 25.2 & 16.0 & 28.1 & 28.6 \\
    Human Level & - & 88.3 & 78.6 & 85.0 & 84.0 & 73.2 & 87.4 & 93.0 & 94.2 & 96.5 & 90.2 & 79.2 & 84.9 & 86.4 & 92.4 & 84.2 & 89.4 & 96.8 & 93.1 & 100.0 & 91.5 & 86.4 \\\hline
    
    \rowcolor[HTML]{ECF4FF} 
    \multicolumn{1}{l|}{\cellcolor[HTML]{ECF4FF}\textit{Proprietary Models}} & & & & & & & & & & & & & & & & & & & & & & \\
    GPT-5.1-251113 & \cellcolor[HTML]{FFCCC9}{3} & 53.4 & \textbf{58.3} & 47.0 & \underline{70.2} & \textbf{53.1} & 46.9 & 32.2 & 27.7 & 44.3 & \textbf{51.6} & 56.7 & \textbf{44.5} & \textbf{42.7} & \underline{53.7} & 42.9 & \textbf{52.2} & \underline{38.2} & 57.8 & 14.0 & \textbf{59.3} & 48.6\\
    Gemini-2.5-Pro & \cellcolor[HTML]{FD6864}{2} & \underline{53.8} & \underline{57.9} & \textbf{55.0} & 57.5 & 47.0 & \underline{50.9} & 33.9 & 41.3 & 51.3 & \underline{50.9} & \textbf{60.8} & 39.5 & 30.1 & 52.0 & 41.4 & 43.4 & \textbf{43.4} & \underline{59.3} & \underline{24.0} & 49.2 & \underline{55.7} \\
    Qwen-3-VL-Plus & 5 & 45.2 & 40.8 & 42.0 & 63.8 & 37.8 & 44.6 & \textbf{47.0} & \underline{41.7} & \textbf{52.2} & 46.5 & \underline{60.0} & 35.3 & 31.8 & 37.1 & \textbf{44.2} & 45.1& 37.1 & 56.7 & 18.0 & 47.5 & 52.1 \\
    Step-1o-turbo-vision & 4 & 51.8 & 55.9 & 42.0 & 41.3 & 46.0 & 48.2 & \underline{39.1} & 35.5 & 47.8 & 45.8 & 56.7 & 39.5 & 34.6 & 52.5 & 41.9 & 45.1 & 37.1 & \textbf{59.5} & 10.0 & \underline{55.9} & \textbf{57.1} \\
    Doubao-seed1.6-251015 & \cellcolor[HTML]{FE0000}{1} & \textbf{54.1} & 57.7 & 43.0 & \textbf{73.4} & \underline{48.4} & \textbf{56.3} & \underline{39.1} & \textbf{51.7} & \textbf{52.2} & 46.9 & 55.0 & \underline{38.7} & \textbf{35.9} & \textbf{54.8} & \underline{43.3} & \underline{47.8} & 36.6 & 58.8 & \textbf{28.0} & \underline{55.9} & \underline{55.7} \\
    Skywork-R1V4-Lite & 6 & 40.1 & 38.6 & \underline{49.0} & 43.6 & 34.3 & 35.8 & 27.0 & 18.2 & 46.9 & 34.6 & 35.8 & 29.4 & 19.6 & 42.9 & 31.6 & 43.4 & 29.6 & 51.0 & 8.0 & 39.0 & 48.6 \\\hline
    
    \rowcolor[HTML]{ECF4FF} 
    \multicolumn{1}{l|}{\cellcolor[HTML]{ECF4FF}\textit{Open-source Models}} & & & & & & & & & & & & & & & & & & & & & & \\
    Qwen3-VL-8B-Instruct & \cellcolor[HTML]{FFCCC9}{3} & 43.1 & 36.1 & 20.0 & 28.7 & 14.1 & 37.6 & 24.4 & 27.3 & 25.7 & \textbf{45.8} & \underline{40.0} & 30.3 & 37.3 & \underline{44.2} & \textbf{37.2} & 38.1 & 34.4 & 49.6 & 22.0 & 44.1 & 30.0 \\
    Qwen3-VL-8B-Thinking & 9 & 39.7 & \underline{41.8} & 22.0 & 41.5 & \underline{35.7} & 36.3 & 28.7 & 22.7 & 20.4 & 39.0 & 35.8 & 21.9 & 36.4 & 39.0 & 20.9 & 36.3 & 30.7 & 42.9 & 18.0 & 35.6 & 22.1 \\
    GLM-4.1V-9B-Base & 5 & 42.6 & 39.7 & 19.0 & 36.2 & 24.4 & 36.2 & 32.2 & 28.5 & 31.0 & 43.0 & \textbf{42.5} & \underline{33.6} & 38.6 & 42.4 & 28.4 & 36.3 & \textbf{37.6} & 51.3 & \underline{32.0} & 44.1 & \underline{32.1} \\
    GLM-4.1V-9B-Thinking & \cellcolor[HTML]{FE0000}{1} & \textbf{44.9} & \textbf{45.8} & \textbf{25.0} & 36.2 & \textbf{40.9} & \underline{42.8} & \textbf{43.5} & 24.8 & \textbf{54.0} & \underline{44.8} & 36.7 & 28.6 & 39.1 & 40.2 & 33.5 & 39.8 & 28.5 & 51.5 & 26.0 & 39.0 & 27.1 \\
    Kimi-VL-A3B-Instruct & 10 & 39.7 & 36.1 & \textbf{25.0} & 12.8 & 23.5 & 33.9 & 27.8 & 24.0 & 45.1 & 36.4 & 35.8 & 24.4 & 35.0 & 43.8 & 28.4 & \underline{40.7} & 35.5 & 49.3 & 10.0 & 39.0 & 23.6 \\
    Kimi-VL-A3B-Thinking & 13 & 36.0 & 32.6 & 18.0 & 34.0 & 27.2 & 31.6 & 22.6 & 19.8 & 27.4 & 34.6 & 25.0 & 26.1 & 21.8 & 39.0 & 25.1 & \textbf{43.4} & 29.0 & 42.1 & 12.0 & 42.4 & 27.1 \\
    MiMo-VL-7B-SFT & 8 & 40.2 & 36.9 & 23.0 & 30.9 & 13.2 & 39.8 & 32.2 & 28.9 & 37.2 & 38.1 & 21.7 & 21.0 & 37.3 & 38.9 & 32.1 & 31.9 & 26.3 & 48.4 & 16.0 & 44.1 & 23.6 \\
    MiMo-VL-7B-RL & 7 & 40.9 & 38.2 & 21.0 & 33.0 & 16.9 & 39.9 & 35.7 & 30.2 & 30.1 & 38.4 & 29.2 & 26.9 & 37.3 & 41.2 & 34.0 & 31.9 & 30.1 & 48.2 & 18.0 & \underline{47.5} & 23.6 \\
    MiniCPM-V-4.5 & \cellcolor[HTML]{FD6864}{2} & \underline{43.9} & 37.1 & 20.0 & 19.2 & 19.3 & \textbf{44.3} & \underline{40.0} & \textbf{34.3} & 33.6 & 42.6 & 36.7 & 26.9 & 35.0 & 43.5 & \underline{35.8} & 38.1 & 31.2 & \underline{52.5} & 26.0 & 37.3 & 27.1 \\
    Ovis2.5-9B & 4 & 42.7 & 40.7 & 20.0 & 30.9 & 24.9 & 41.1 & 36.5 & \underline{33.1} & 38.1 & 39.3 & 20.8 & 22.7 & \underline{45.5} & 40.9 & 24.7 & 39.8 & 31.2 & \textbf{53.2} & \textbf{40.0} & \underline{47.5} & 28.6 \\
    LLaVA-NeXT-Video-7B & 16 & 28.3 & 32.6 & \textbf{25.0} & \underline{42.6} & 23.0 & 21.5 & 25.2 & 25.2 & 8.0 & 25.9 & 32.5 & 18.5 & 23.2 & 26.3 & 23.7 & 26.6 & 23.1 & 36.6 & 20.0 & 35.6 & 29.3 \\
    LLaVA-Onevision-7B & 6 & 42.3 & 37.4 & 22.0 & 28.7 & 27.2 & 34.8 & 28.7 & 25.2 & \underline{48.7} & 43.2 & 30.8 & 28.6 & \textbf{50.5} & \textbf{45.6} & 24.7 & 37.2 & 37.1 & 49.6 & 20.0 & 30.5 & 29.3 \\
    InternVL2.5-8B & 12 & 38.7 & 36.1 & 16.0 & \underline{45.7} & 14.1 & 31.7 & 17.4 & 18.6 & 44.3 & 37.2 & 34.2 & \underline{33.6} & 33.2 & 40.4 & 24.2 & \underline{40.7} & \textbf{37.6} & 48.2 & 10.0 & \textbf{50.9} & 25.0 \\
    Skywork-VL-Reward-7B & 14 & 33.2 & 36.8 & 5.0 & 41.5 & 28.2 & 34.9 & 27.0 & 24.4 & 36.3 & 22.8 & 2.5 & 21.0 & 21.8 & 32.7 & 21.9 & 23.9 & 19.4 & 44.6 & 28.0 & 35.6 & 19.3 \\
    Molmo-7B-D-0924 & 11 & 38.7 & 33.3 & 17.0 & 25.5 & \underline{30.5} & 33.5 & 31.3 & 20.7 & 30.1 & 36.8 & 28.3 & \textbf{34.5} & 32.3 & 41.8 & 26.5 & 33.6 & 28.0 & 48.2 & 26.0 & 44.1 & \textbf{33.6} \\
    Phi-4-multimodal-instruct & 15 & 32.0 & 31.8 & 23.0 & \textbf{46.8} & 16.4 & 29.0 & 33.9 & 17.8 & 40.7 & 27.3 & 11.7 & 16.8 & 25.0 & 32.0 & 28.4 & 15.0 & 24.7 & 42.0 & 18.0 & 32.2 & 17.1 \\\hline
    
    \rowcolor[HTML]{ECF4FF} 
    \multicolumn{1}{l|}{\cellcolor[HTML]{ECF4FF}\textit{Spatial Models}} & & & & & & & & & & & & & & & & & & & & & & \\
    Spatial-SSRL-4B & \cellcolor[HTML]{FE0000}{1} & \textbf{39.8} & \textbf{39.2} & \underline{22.0} & 35.1 & 15.0 & 31.2 & \underline{31.3} & \underline{26.5} & 21.2 & \textbf{41.6} & \underline{30.0} & \textbf{31.1} & \textbf{47.3} & 37.8 & \textbf{30.2} & \underline{35.4} & 24.7 & \underline{48.0} & \textbf{34.0} & \textbf{49.2} & 27.1 \\
    SpaceOm-4B & \cellcolor[HTML]{FD6864}{2} & \underline{38.9} & \underline{38.6} & \textbf{25.0} & \textbf{40.4} & \textbf{23.5} & \textbf{34.3} & \textbf{33.9} & 23.1 & \underline{48.7} & \underline{37.1} & \textbf{35.0} & 19.3 & \underline{33.2} & \underline{37.9} & 26.5 & \textbf{36.3} & \underline{34.9} & 47.2 & 8.0 & \underline{45.8} & \textbf{35.0} \\
    SpaceThinker-3B & \cellcolor[HTML]{FFCCC9}{3} & 38.7 & 35.8 & 21.0 & \underline{38.3} & \underline{19.3} & \underline{34.2} & 25.2 & \textbf{29.8} & \textbf{55.8} & 35.6 & 25.0 & \underline{29.4} & 30.9 & \textbf{40.5} & \underline{29.8} & 33.6 & \textbf{36.0} & \textbf{48.5} & \underline{18.0} & 33.9 & \underline{33.6} \\
    \hhline{=======================} 
    
    \rowcolor[HTML]{ECF4FF} 
    \multicolumn{1}{l|}{\cellcolor[HTML]{ECF4FF}\textit{Fine-Tuning: Test set}} & & & & & & & & & & & & & & & & & & & & & & \\
    Qwen3-VL-2B (before) & 9 & 30.4 & 26.4 & 30.0 & 10.0 & 22.7 & 23.4 & 25.0 & 12.0 & 33.3 & 37.1 & 50.0 & \underline{41.7} & 31.8 & 25.4 & 13.6 & 16.7 & 26.3 & 36.7 & 20.0 & 50.0 & 28.6 \\
    CityBot-2B (CoT) & 4 & 60.2 & 61.5 & 40.0 & \textbf{70.0} & 59.1 & \textbf{62.8} & 48.3 & \textbf{56.0} & \underline{83.3} & 60.1 & \textbf{75.0} & \underline{41.7} & 45.5 & 50.9 & \underline{31.8} & 38.3 & 31.6 & \underline{67.4} & \underline{60.0} & 50.0 & 57.1 \\
    CityBot-2B (w/o CoT) & 6 & 55.8 & 58.2 & 40.0 & 50.0 & 68.2 & 50.0 & 41.7 & 28.0 & 75.0 & 57.3 & 66.7 & \underline{41.7} & 45.5 & 46.1 & 18.2 & 33.3 & 42.1 & \underline{67.4} & 40.0 & 50.0 & 60.0 \\
    Qwen3-VL-4B (before) & 8 & 36.6 & 38.5 & 20.0 & 60.0 & 40.9 & 27.7 & 41.7 & 28.0 & 12.5 & 37.8 & 50.0 & 25.0 & 36.4 & 39.2 & 27.3 & \underline{41.7} & 15.8 & 38.8 & 40.0 & 33.3 & 28.6 \\
    CityBot-4B (CoT) & \cellcolor[HTML]{FD6864}{2} & \underline{61.0} & \textbf{67.0} & \textbf{50.0} & \textbf{70.0} & \underline{72.7} & \underline{59.6} & \textbf{50.0} & 48.0 & 66.7 & \textbf{58.1} & 66.7 & \underline{41.7} & \underline{50.0} & \underline{54.9} & 27.3 & \underline{41.7} & \textbf{68.4} & \underline{67.4} & \textbf{80.0} & 50.0 & \textbf{71.4} \\
    CityBot-4B (w/o CoT) & \cellcolor[HTML]{FFCCC9}{3} & 60.4 & 62.6 & 34.3 & 50.0 & 68.2 & 53.2 & 33.3 & 28.0 & 75.0 & \underline{58.0} & 66.7 & \underline{41.7} & 45.5 & \textbf{59.8} & \underline{31.8} & \underline{41.7} & \underline{52.6} & \textbf{69.4} & 40.0 & \textbf{83.3} & 64.3 \\
    Qwen3-VL-8B (before) & 7 & 37.1 & 40.7 & 20.0 & 60.0 & 45.5 & 26.6 & 8.3 & 16.0 & 58.3 & 42.0 & 66.7 & 33.3 & 31.8 & 38.2 & 22.7 & \underline{41.7} & 26.3 & 35.7 & 40.0 & 50.0 & 57.1 \\
    CityBot-8B (CoT) & \cellcolor[HTML]{FE0000}{1} & \textbf{61.4} & \underline{64.8} & \textbf{50.0} & \textbf{70.0} & \textbf{77.3} & \textbf{62.8} & 41.7 & \underline{52.0} & \textbf{91.7} & \underline{58.0} & \textbf{75.0} & \textbf{50.0} & \textbf{54.6} & 52.9 & \textbf{40.9} & \textbf{50.0} & 36.8 & 64.8 & \underline{60.0} & \underline{66.7} & \textbf{71.4} \\
    CityBot-8B (w/o CoT) & 5 & 57.8 & 58.2 & 40.0 & 50.0 & 68.2 & 54.3 & \textbf{50.0} & 32.0 & \underline{83.3} & 57.3 & \textbf{75.0} & 33.3 & 45.5 & 53.9 & \underline{31.8} & 25.0 & \underline{52.6} & 65.3 & 20.0 & \underline{66.7} & 50.0 \\
    \hline
\end{tabularx}
\vspace{-10pt}
\end{table*}

Our main observations based on the results shown in Table~\ref{tab:complete_comparison} are summarized as follows:


\textbf{Limited cross-view spatial intelligence across current VLMs.}
We find that \emph{both} proprietary and open-source VLMs perform poorly on CityCube, indicating that CvSI remains largely unsolved by existing model families. The best-performing proprietary model achieves only $54.1\%$ accuracy, exceeding the strongest open-source model by $9.2\%$, yet still falling far short of human performance ($-34.2\%$). This consistent gap suggests that the difficulty lies not in model architecture or scale alone, but in the fundamental challenge of multi-view spatial reasoning posed by CityCube.
\textbf{Effectiveness of fine-tuning on CityCube.}
Fine-tuning on CityCube consistently improves model performance across all scales. Additionally, training with human annotations (e.g., CityBot-4B and -8B with CoT) yields further gains ($+0.6\%$ to $+3.6\%$), indicating the benefit of reasoning guidance. Notably, the fine-tuned 2B model already surpasses strong proprietary baselines, highlighting the advantage of the benchmark.

\textbf{Reasoning-oriented VLMs struggle with spatial tasks.}
Despite their success in math and coding, reasoning-oriented models (e.g., Qwen3-VL-8B-Thinking, Kimi-VL-A3B-Thinking) show no consistent advantage over non-reasoning models on multi-view spatial tasks such as \textit{PT}. This suggests that generic reasoning supervision alone is insufficient to induce strong spatial reasoning in urban environments. We hypothesize that such reasoning requires explicit modeling of view-dependent geometry and spatial transformations, which is not encouraged by current post-training strategies.

\textbf{CityCube reveals limitations in existing benchmarks.}
General-purpose spatial models like SpaceOM perform sub-optimally on our urban embodied scenarios. This finding validates the shortcomings of existing benchmarks and emphasizes the unique value of CityCube in evaluating the distinct challenges of CvSI, which are not covered by standard visual question answering evaluations.

\textbf{Different spatial dimensions exhibit uneven difficulty.}
Models perform relatively well on \textit{WK} and \textit{CR} tasks. These tasks mainly rely on semantic understanding and logical inference. In contrast, performance drops significantly on \textit{PT}, \textit{SR}, and \textit{MR}. They require precise geometric reasoning and viewpoint transformation across multiple views. The results indicate that current VLMs favor semantic priors over robust spatial representations.

\subsection{Task Correlation Analysis}
\begin{figure}
\centering
\includegraphics[width=\linewidth]{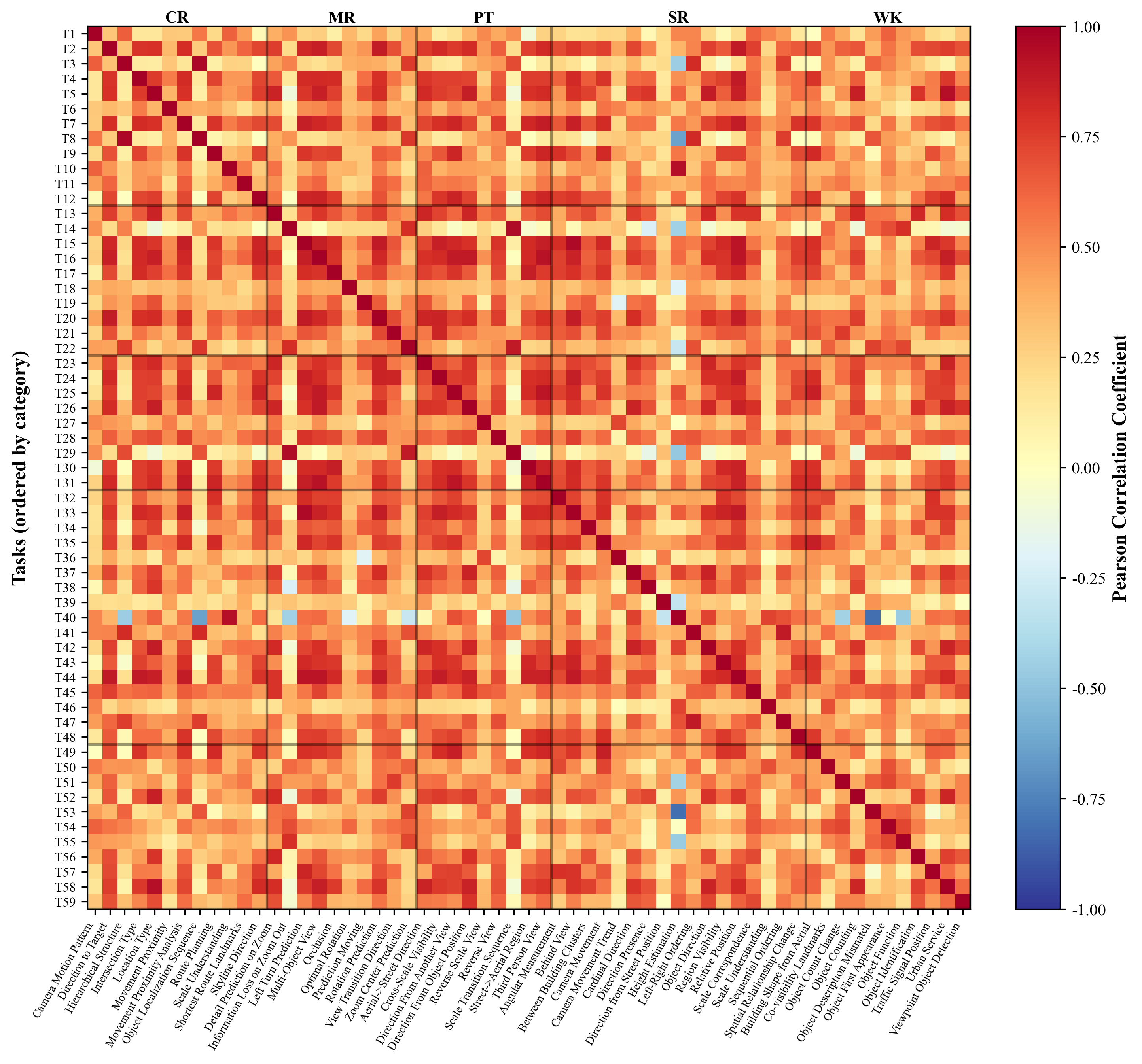}
\caption{Task correlation matrix. Each axis corresponds to the 59 tasks after classification, and color intensity indicates the strength of correlations.}
\label{fig:task_corr}
\end{figure}

We posit that tasks requiring similar cognitive capabilities will elicit correlated model performance. To analyze the structural relationships among CvSI tasks, we compute the Pearson correlation matrix over $59$ tasks evaluated on $25$ baseline VLMs. As shown in Fig.~\ref{fig:task_corr}, we report three key findings:

\textbf{Strong correlations across task categories.}
At the dimension level, we observe generally substantial correlations across the five CvSI categories, indicating that spatial intelligence is not naturally decomposed into independent modules. Among all pairs, \textit{MR} and \textit{PT} exhibit the highest inter-dimension correlation ($r=0.536$), suggesting a shared reliance on the underlying cognitive mechanism.

\textbf{Dense correlations across different tasks.}
At a finer granularity, task-level analysis reveals dense cross-category correlations, with $81.4\%$ of high-correlation pairs spanning different CvSI dimensions. For example, \textit{Behind View} (T33, SR) and \textit{Left-turn Prediction} (T15, CR) exhibit near-perfect correlation ($r=0.954$), indicating that panoramic spatial perception is tightly coupled with dynamic viewpoint reasoning. Similarly, strong associations are observed between cross-scale semantic tasks such as \textit{Hierarchical Structure} (T3, WK) and \textit{Object Localization Sequence} (T8, CR) ($r=0.965$), reflecting synchronized semantic–geometric reasoning under dynamic scale changes.

\textbf{Metric estimation as a weakly correlated capability.}
Metric estimation shows weaker coupling with other tasks. For example, \textit{Height Estimation} (T40) shows negligible correlation with most other tasks, suggesting that precise metric reasoning constitutes a distinct capability, weakly linked to view-dependent or semantic spatial reasoning processes.

\subsection{The VLMs Error Analysis}
Through a case-by-case qualitative investigation (as depicted in Appendix \ref{sec:error_details}), we identify four primary failure modes of current VLMs in urban spatial reasoning:

\textbf{Limited sensitivity to small-scale urban entities.}
In complex urban environments, models frequently fail to recognize small-scale urban entities or entirely overlook critical landmarks. This issue is particularly pronounced in scenes with high semantic density, where numerous visually similar objects compete for attention.

\textbf{Failures in egocentric spatial reasoning.}
Models struggle to correctly predict changes in relative position, orientation, and visibility when the egocentric viewpoint shifts. This failure in ``mental rotation" indicates a lack of a reliable internal spatial representation.

\textbf{Insufficient cross-view consistency.}
Models exhibit difficulty in establishing correct correspondences across perspectives. For instance, models often mismatch a building in street imagery with an unrelated structure in aerial views, leading to erroneous reasoning about spatial relationships and world knowledge.

\textbf{Misinterpretation of motion and scale dynamics.}
Models struggle to interpret the camera movements (e.g., forward/backward translation or directional turns) and the resulting dynamic scale changes. Such failures in motion understanding directly impede the ability to perform complex scene reconstruction and navigation-related reasoning.
\subsection{Human-AI Difficulty Bias Analysis}
\begin{figure}
\centering
\includegraphics[width=.95\linewidth]{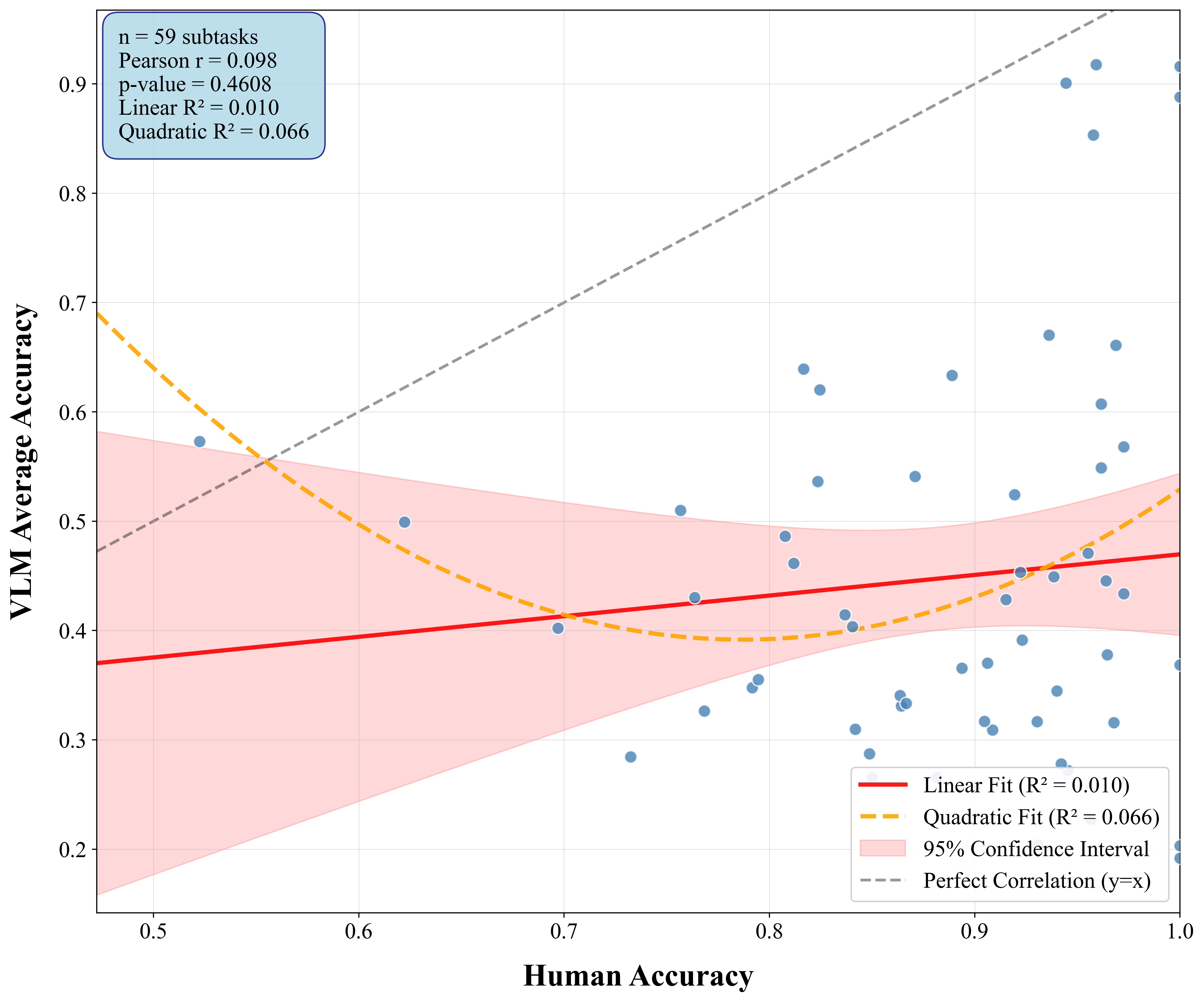}
\caption{Human vs AI task correlation. The scatter plots illustrate the performance correlation across 59 tasks, where each individual point represents a specific task.}
\label{fig:humanAI_corr}
\end{figure}
To examine whether the difficulty of spatial tasks for humans aligns with VLM performance, we analyze the correlation between human baseline accuracy and the average performance of evaluated VLMs across 59 subtasks. Fig.~\ref{fig:humanAI_corr} reveals a Pearson correlation coefficient of $r = 0.098$ with a p-value of $0.4608$. This extremely low correlation ($R^2 = 0.010$) indicates that tasks found difficult by VLMs do not necessarily pose a challenge for humans, and vice versa. Notably, the AI performance distribution is highly dispersed (range: $[19.2\%, 91.8\%]$), suggesting that VLMs might be merely sensitive to low-level visual features rather than spatial mental modeling. This divergence confirms that CityCube captures unique spatial challenges that are non-trivial for current model architectures, despite being intuitive for humans.

%% file: sec/5_Conclusion.tex
\section{Conclusion}
In this paper, we presented \textbf{CityCube}, a comprehensive benchmark specifically designed to evaluate CvSI of VLMs in urban environments. CityCube encompasses 59 tasks across 5 cognitive categories, supported by a large-scale collection of images from dynamic viewpoints and diverse orientations. The resulting dataset contains 5,022 multiple-choice questions (MCQs), each rigorously annotated and verified by humans. 

Our extensive evaluation of 33 VLMs reveals that CvSI remains extremely challenging, even for very large-scale models, ``thinking'' models and specialized spatial models.
While our fine-tuned \textbf{CityBot} models (based on 2B, 4B and 8B backbones) outperform leading proprietary models, a substantial gap to human-level spatial cognition persists.
We hope that CityCube can serve as a foundation for future studies on spatially grounded learning paradigms and as a diagnostic tool for developing next-generation VLMs with stronger urban spatial intelligence.


\section{Limitations}
This work has several limitations. First, although CityCube covers diverse cross-view spatial tasks, we do not explicitly isolate perspective-induced cognitive biases, such as viewpoint asymmetry and reference-frame ambiguity, which may systematically influence spatial reasoning performance. Second, CityCube is constructed in a simulated urban environment; the impact of Sim-to-Real transfer and domain gaps is not evaluated in this study. Third, our analysis focuses on task-level performance and does not probe internal representations or multi-view fusion mechanisms, limiting interpretability of model failure modes. Finally, while we hypothesize the importance of explicit spatial supervision, we do not implement or validate dedicated post-training strategies (e.g., spatial Chain-of-Thought) in this work.
Future work will address these issues by analyzing viewpoint biases, extending evaluation to Sim2real settings, and exploring spatially grounded architectural and post-training designs.

\section{Ethics Statement}
This research exclusively utilizes publicly available datasets, programs, and pre-trained models. All data annotation was conducted with informed consent from participants, and the datasets do not contain information that could compromise individual privacy or public safety. All procedures strictly adhere to the guidelines established by the ACL Code of Ethics. Therefore, this work does not raise any ethical concerns.

%% file: sec/X_suppl.tex
\clearpage
\setcounter{page}{1}
\section{Appendix}
\label{sec:appendix}
In the supplementary materials, we provide the following:
\begin{itemize}
    \item Details of data collection, processing, automatic generation and refinement of \textbf{CityCube} Benchmark (Sec \ref{sec:pipeline}).
    \item Details of reproduction, including code for QA generation, model training scripts, image datasets, and the detail model information (Sec \ref{sec:exp_details}).
    \item Details of VLMs error and their corresponding reason (Sec \ref{sec:error_details}).
    \item Details of experiment setup, including training hyperparameter, dataset settings and human evaluation UI interface (Sec \ref{sec:humaneval}).
    \item Further discussion on relative research, especially in 3D QA (Sec \ref{sec:furtherdiscuss}).
\end{itemize}

\section{Dataset Generation Pipeline}
\label{sec:pipeline}
\begin{figure*}
\centering
\includegraphics[width=.95\linewidth]{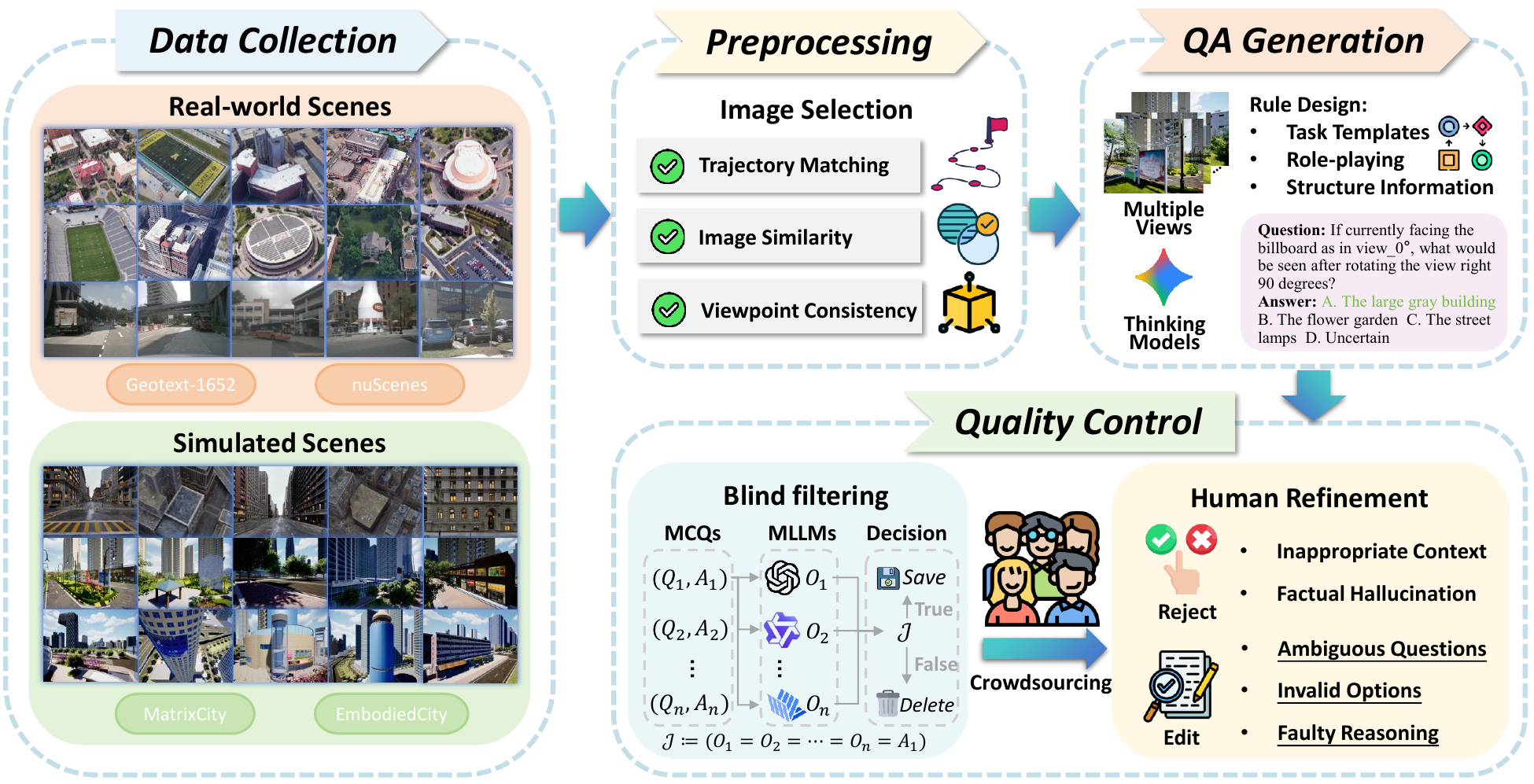}
\caption{Illustration of the CityCube-Bench construction pipeline. Images are collected from diverse real-world datasets and urban simulators; relevant image sets are carefully selected; complex QA tasks and detailed reasoning processes are annotated through human–AI collaboration; and all data undergo rigorous quality control.}
\label{fig:pipeline}
\end{figure*}

\subsection{Dataset Visualization}
\label{subsec:dataset_visualize}
\begin{figure}
    \centering
    \includegraphics[width=\linewidth]{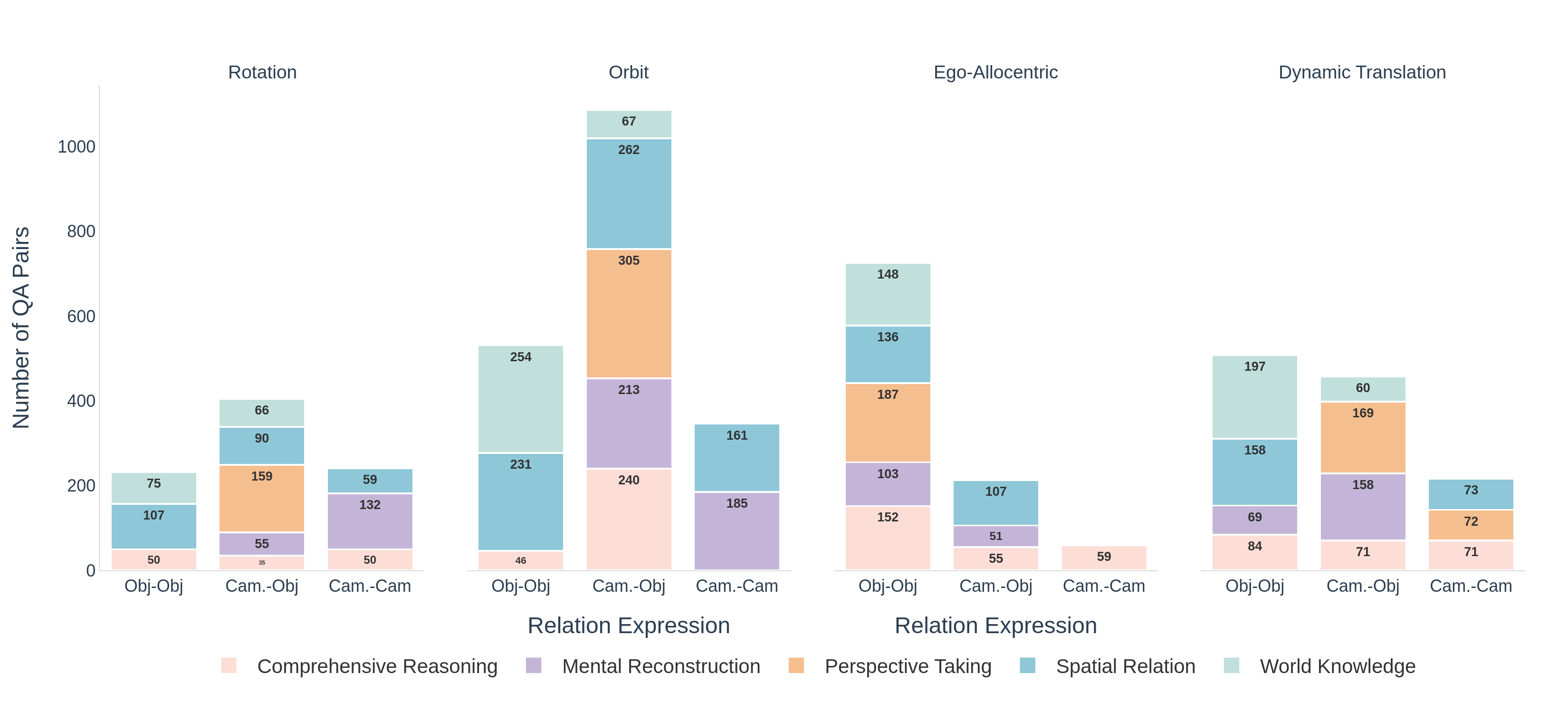}
    \caption{Histogram of different behaviors and relation expressions}
    \label{fig:stacked_bar}
\end{figure}

\begin{figure}
\centering
\includegraphics[width=\linewidth]{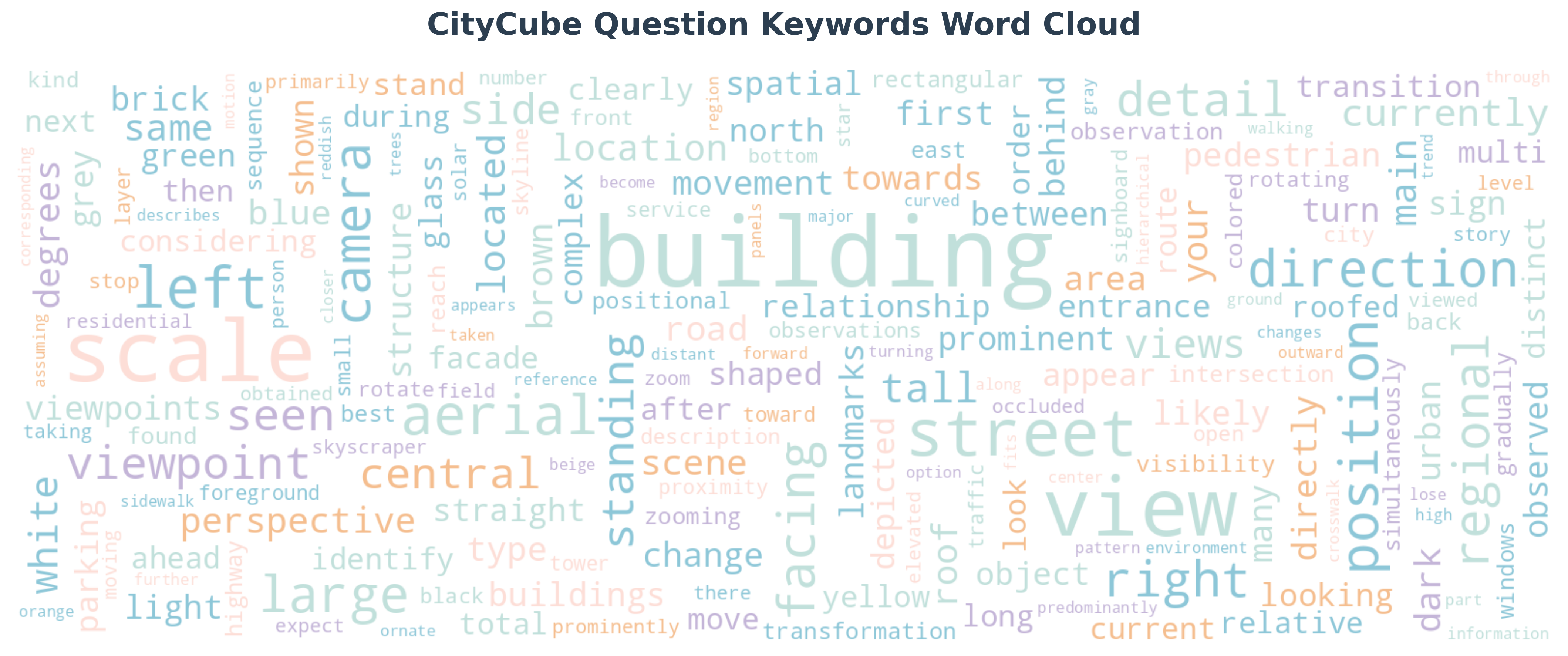}
\caption{Word cloud of CityCube QA dataset.}
\label{fig:text_cloud}
\end{figure}

We analyzed the distribution of spatial reference frames and spatial task frequencies associated with each observation behavior, with the results summarized in the stacked bar chart in Fig. \ref{fig:stacked_bar}.
The training and test sets contain approximately 4.5k and 0.5k questions, respectively. We construct each split using a stratified sampling strategy, ensuring that the proportion of each task type remains consistent with the overall QA distribution of the dataset. We further present word cloud visualizations of the textual content in the training set, as illustrated in Fig.~\ref{fig:text_cloud}.

\subsection{Details of Image Acquisition}
\label{subsec:images}

Our sources span both real-world environments and photorealistic virtual scenes. We leverage real-world sensor datasets—GeoText-1652 \cite{chu2024towards} and nuScenes \cite{caesar2020nuscenes}, to acquire raw data for perspectives involving camera behavior primitives of Rotation and Orbit. We further augment the benchmark with extensive first-person to third-person camera motion imagery derived from embodied vision, which are based on virtual environments—EmbodiedCity \cite{gao2024embodiedcity} and MatrixCity \cite{li2023matrixcity}.

The core datasets are selected for their complementary strengths in providing diverse, high-quality visual data while adhering to rigorous data ethics standards. Key details are outlined below:
\begin{itemize}
    \item \textbf{nuScenes \cite{caesar2020nuscenes}}: This multimodal autonomous driving dataset provides 1,000 large-scale scenes from real-world urban environments (Boston and Singapore). Its synchronized data from six cameras, LiDAR, and radar, along with comprehensive 3D annotations and precise calibration, offers a rich, real-world foundation for studying multi-view geometry and complex camera motions.
    
    \item \textbf{GeoText-1652 \cite{chu2024towards}}: This benchmark extends real-world imagery with spatial language annotations. Based on the established University-1652 image set, it provides high-resolution ground-level and aerial images. Its key contribution to our work is the precise 360-degree image correspondence the aerial scan of urban buildings, facilitating tasks that require spatial and linguistic grounding.
    
    \item \textbf{MatrixCity \cite{li2023matrixcity}}: A large-scale photorealistic synthetic dataset built with Unreal Engine 5. It provides over 500,000 street-view and aerial images with pixel-perfect ground-truth information (e.g., camera poses, depth, normal maps) and full control over environmental conditions (weather, lighting). It serves as a primary, privacy-safe source for generating diverse first-person camera motion trajectories.
    
    \item \textbf{EmbodiedCity \cite{gao2024embodiedcity}}: This benchmark platform supports embodied AI agents in city-scale environments. It enables the generation of extensive, realistic first-person visual experience data for navigation and interaction tasks, perfectly aligning with our need for embodied, egocentric visual data in complex virtual urban settings.
\end{itemize}

\noindent \textbf{Privacy and Ethical Compliance Statement.} We strictly adhere to data privacy and ethical research standards. All real-world data utilized in this study (nuScenes and GeoText-1652) are sourced from established, publicly released academic datasets that have undergone formal anonymization and curation processes, involving no collection of personal identifiers. Crucially, a significant portion of our training data—particularly for complex, embodied first-person perspectives—is generated from fully synthetic environments (MatrixCity and EmbodiedCity). This strategy inherently eliminates any risk associated with the privacy of individuals, as no real human subjects or private spaces are involved. All data is used strictly for non-commercial academic research in full compliance with their respective licenses.

\subsection{Details of Multi-view Images Processing}
\label{subsec:images process}
\subsubsection{nuScenes}
\label{app:data_processing}
Due to the high temporal redundancy in continuous driving sequences, directly using all frames would introduce substantial duplication and bias the distribution toward near-identical views. To address this issue, we adopt a two-stage frame reduction strategy: \emph{temporal subsampling} followed by \emph{appearance-based deduplication}. Specifically, frames are first sampled at a fixed temporal interval to reduce redundancy at the sequence level. Then, a lightweight visual similarity test is applied to further filter near-duplicate frames.

For each candidate timestep, synchronized images from six surrounding cameras are aggregated into a compact multi-view representation. The average pixel-wise absolute difference between consecutive aggregated views is used as a proxy for visual change. Frames with appearance variation below a predefined threshold are discarded. This procedure preserves scene diversity while significantly reducing redundant observations. The retained frames are stored as structured multi-view samples, each associated with metadata describing scene identity and camera configuration.

\begin{algorithm}[t]
\caption{Multi-view Frame Temporal Sampling and Deduplication for nuScenes}
\label{alg:multiview_export}
\KwIn{
Driving dataset $\mathcal{D}$ with scenes and synchronized multi-camera frames; \\
Temporal step size $s$; similarity threshold $\tau$; optional per-scene limit $M$
}
\KwOut{
A set of exported multi-view frame groups with metadata
}

\ForEach{scene $\mathcal{S}$ in $\mathcal{D}$}{
    Retrieve ordered frame list $\{\mathbf{F}_1, \mathbf{F}_2, \dots, \mathbf{F}_N\}$\;
    Initialize $\mathbf{A}_{\text{prev}} \leftarrow \varnothing$, counter $k \leftarrow 0$\;

    \For{$i = 1, 1+s, \dots, N$}{
        \If{$M$ is set \textbf{and} $k \ge M$}{
            \textbf{break}
        }

        Let $\mathbf{F}_i = \{I_i^{(1)}, \dots, I_i^{(6)}\}$ be six synchronized camera images\;
        \If{any image in $\mathbf{F}_i$ is missing}{
            \textbf{continue}
        }

        Construct concatenate view $\mathbf{A}_i \leftarrow \textsc{Concate}(\mathbf{F}_i)$\;

        \If{$\mathbf{A}_{\text{prev}} \neq \varnothing$}{
            Compute appearance difference
            \[
            d \leftarrow \frac{1}{|\mathbf{A}_i|} 
            \sum \left| \mathbf{A}_i - \mathbf{A}_{\text{prev}} \right|
            \]
            \If{$d < \tau$}{
                \textbf{continue}
            }
        }

        Export all images in $\mathbf{F}_i$ as one multi-view sample\;
        Save associated metadata (scene ID, frame index, camera list)\;
        $\mathbf{A}_{\text{prev}} \leftarrow \mathbf{A}_i$\;
        $k \leftarrow k + 1$\;
    }
}
\end{algorithm}

\subsubsection{Geotext-1652}
Unlike driving datasets with strong temporal continuity, images in GeoText-1652 are organized as viewpoint sequences centered around prominent landmarks, exhibiting systematic variations in camera distance and altitude.

To obtain representative landmark-centric multi-view observations while avoiding redundant samples, we adopt a rule-based sampling strategy with manually designed intervals.
Specifically, for each landmark sequence, we extract two complementary subsets:
(i) \emph{orbit views}, which capture appearance changes under small viewpoint variations at similar altitudes,
and (ii) \emph{spiral views}, which emphasize significant altitude and scale changes.
These two subsets jointly approximate local and global perceptual variations around a landmark.

Given an ordered image sequence, orbit views are sampled with a short interval to retain fine-grained viewpoint diversity, while spiral views are sampled with a larger interval to highlight elevation changes.
To further control redundancy and balance the dataset, we impose an upper bound on the number of retained images per subset.
The selected images are exported into separate directories according to their view type, preserving the original filenames for traceability.
This strategy yields compact yet diverse multi-view observations suitable for landmark-centric perception and reasoning tasks.

\begin{algorithm}[t]
\caption{Landmark-centric Multi-view Sampling for GeoText-1652}
\label{alg:geotext_sampling}

\KwIn{
Image folders $\{\mathcal{F}_1, \dots, \mathcal{F}_K\}$; \\
orbit sampling step $s_o$; spiral sampling step $s_s$; \\
maximum samples per type $M$
}
\KwOut{
Two subsets per folder: orbit views and spiral views
}

\ForEach{folder $\mathcal{F}$}{
    Load ordered image sequence $\{I_1, I_2, \dots, I_N\}$\;

    Initialize orbit index set $\mathcal{I}_o \leftarrow \varnothing$\;
    Initialize spiral index set $\mathcal{I}_s \leftarrow \varnothing$\;

    \For{$i = 1, 1+s_o, \dots, N$}{
        Append $i$ to $\mathcal{I}_o$\;
    }

    \For{$i = 1, 1+s_s, \dots, N$}{
        Append $i$ to $\mathcal{I}_s$\;
    }

    \If{$N \notin \mathcal{I}_o$}{
        Append $N$ to $\mathcal{I}_o$\;
    }
    \If{$N \notin \mathcal{I}_s$}{
        Append $N$ to $\mathcal{I}_s$\;
    }

    Truncate $\mathcal{I}_o$ and $\mathcal{I}_s$ to at most $M$ elements\;

    \ForEach{$i \in \mathcal{I}_o$}{
        Export image $I_i$ to orbit subset\;
    }

    \ForEach{$i \in \mathcal{I}_s$}{
        Export image $I_i$ to spiral subset\;
    }
}
\end{algorithm}

\begin{figure}
\centering
\includegraphics[width=\linewidth]{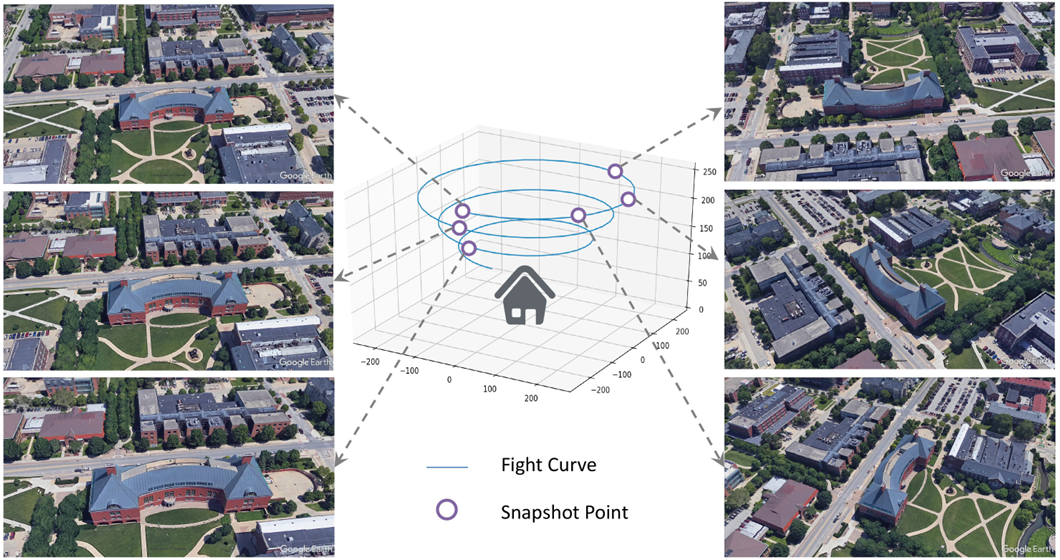}
\caption{A data example of viewpoint dynamic in Geotext-1652.}
\label{fig:geotext}
\end{figure}

\subsubsection{Cross-view Geometric Pairing in MatrixCity}
MatrixCity is a large-scale virtual city dataset providing synchronized aerial and street-level imagery with precise camera poses.
Unlike real-world datasets where camera geometry may be noisy or partially missing, MatrixCity offers accurate extrinsic parameters, enabling explicit geometric reasoning between ego-centric (street) and allocentric (aerial) views.

To construct reliable aerial-street image pairs, we perform geometry-aware cross-view matching based on camera poses and viewing configurations (as depicted in \ref{alg:matrixcity_pairing}).
Given a street-level image, candidate aerial views are filtered and scored through a sequence of geometric constraints.
Only image pairs that satisfy both spatial overlap and viewpoint consistency are retained, while ambiguous or degenerate cases are manually filtered out.
This process ensures strong correspondence between ego-centric observations and their allocentric counterparts.

Specifically, the pairing procedure relies on five core geometric factors:
(1) horizontal Euclidean distance between camera positions in the ground plane;
(2) height difference between aerial and street cameras;
(3) overlap of projected viewing areas on the ground;
(4) consistency of viewing orientation in the horizontal plane, and
(5) ground-plane projection of camera viewing rays.
These factors are applied sequentially to prune invalid candidates and to compute a final matching score for pair selection.

\begin{algorithm}[htp]
\caption{Geometry-aware Aerial--Street Pairing for MatrixCity}
\label{alg:matrixcity_pairing}

\KwIn{
Street images $\mathcal{S}=\{S_i\}$ with poses; \\
Aerial images $\mathcal{A}=\{A_j\}$ with poses; \\
Height range $[h_{\min}, h_{\max}]$; distance threshold $d_{\max}$; \\
Orientation threshold $\theta_{\max}$
}
\KwOut{
Paired aerial--street image set $\mathcal{P}$
}

Initialize $\mathcal{P} \leftarrow \varnothing$\;

\ForEach{street image $S_i \in \mathcal{S}$}{
    Extract street camera position $\mathbf{p}_s$ and orientation $\mathbf{o}_s$\;

    \If{$S_i$ violates diversity constraints}{
        continue\;
    }

    Compute ground projection $(\mathbf{g}_s, r_s)$ from $(\mathbf{p}_s, \mathbf{o}_s)$\;

    Initialize best match $A^\star \leftarrow \varnothing$, best score $c^\star \leftarrow \infty$\;

    \ForEach{aerial image $A_j \in \mathcal{A}$}{
        Extract aerial camera position $\mathbf{p}_a$ and orientation $\mathbf{o}_a$\;

        \tcp{Height filtering}
        \If{$(\mathbf{p}_a^z - \mathbf{p}_s^z) \notin [h_{\min}, h_{\max}]$}{
            continue\;
        }

        \tcp{Horizontal distance filtering}
        Compute $d_h \leftarrow \lVert \mathbf{p}_a^{xy} - \mathbf{p}_s^{xy} \rVert_2$\;
        \If{$d_h > d_{\max}$}{
            continue\;
        }

        \tcp{Ground projection and viewing overlap}
        Compute ground projection $(\mathbf{g}_a, r_a)$ from $(\mathbf{p}_a, \mathbf{o}_a)$\;
        Compute viewing center distance $d_g \leftarrow \lVert \mathbf{g}_a^{xy} - \mathbf{g}_s^{xy} \rVert_2$\;
        \If{$d_g > r_a + r_s$}{
            continue\;
        }

        \tcp{Cross-view orientation consistency}
        Compute horizontal orientation angle $\theta \leftarrow \angle(\mathbf{o}_a^{xy}, \mathbf{o}_s^{xy})$\;
        \If{$\theta > \theta_{\max}$}{
            continue\;
        }

        \tcp{Matching score}
        Compute score $c \leftarrow \alpha d_h + \beta d_g + \gamma (\mathbf{p}_a^z - \mathbf{p}_s^z)$\;

        \If{$c < c^\star$}{
            $c^\star \leftarrow c$, $A^\star \leftarrow A_j$\;
        }
    }

    \If{$A^\star \neq \varnothing$}{
        Add pair $(S_i, A^\star)$ to $\mathcal{P}$\;
        Update diversity state\;
    }
}
\end{algorithm}

We emphasize that the image data does not assume perfect viewpoint control in real-world data. Instead, it focuses on constructing view sets with consistent scene identity and interpretable viewpoint variation, which is sufficient for evaluating cross-view spatial reasoning.

\subsection{Details of Manually Collected Data}
\label{subsec:manual_images}
Cross-scale perception refers to partitioning the urban environment into multiple levels of spatial granularity, for example, progressively reducing the scale from a block-level viewpoint to that of an individual building (or facility), and further down to the object level, such as billboards, trees, or sculptures. With the assistance of experienced simulator operators in our team, we collected cross-scale image sequences under large-range urban motions, resulting in a total of 246 sequences.

\begin{figure}[t]
    \centering
    \begin{subfigure}[t]{\linewidth}
        \centering
        \includegraphics[width=\linewidth]{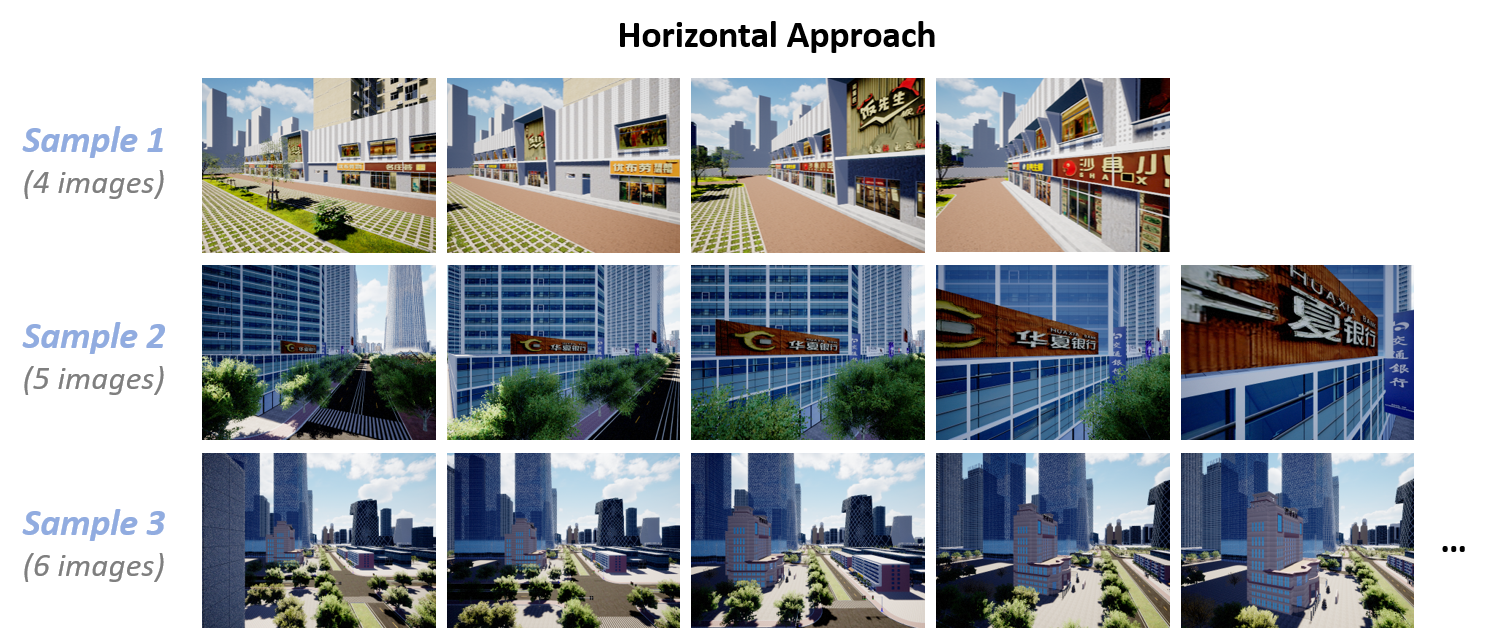}
        \caption{Multi-scale views in horizontal approach during dynamic translation.}
    \end{subfigure}

    \vspace{0.5em}

    \begin{subfigure}[t]{\linewidth}
        \centering
        \includegraphics[width=\linewidth]{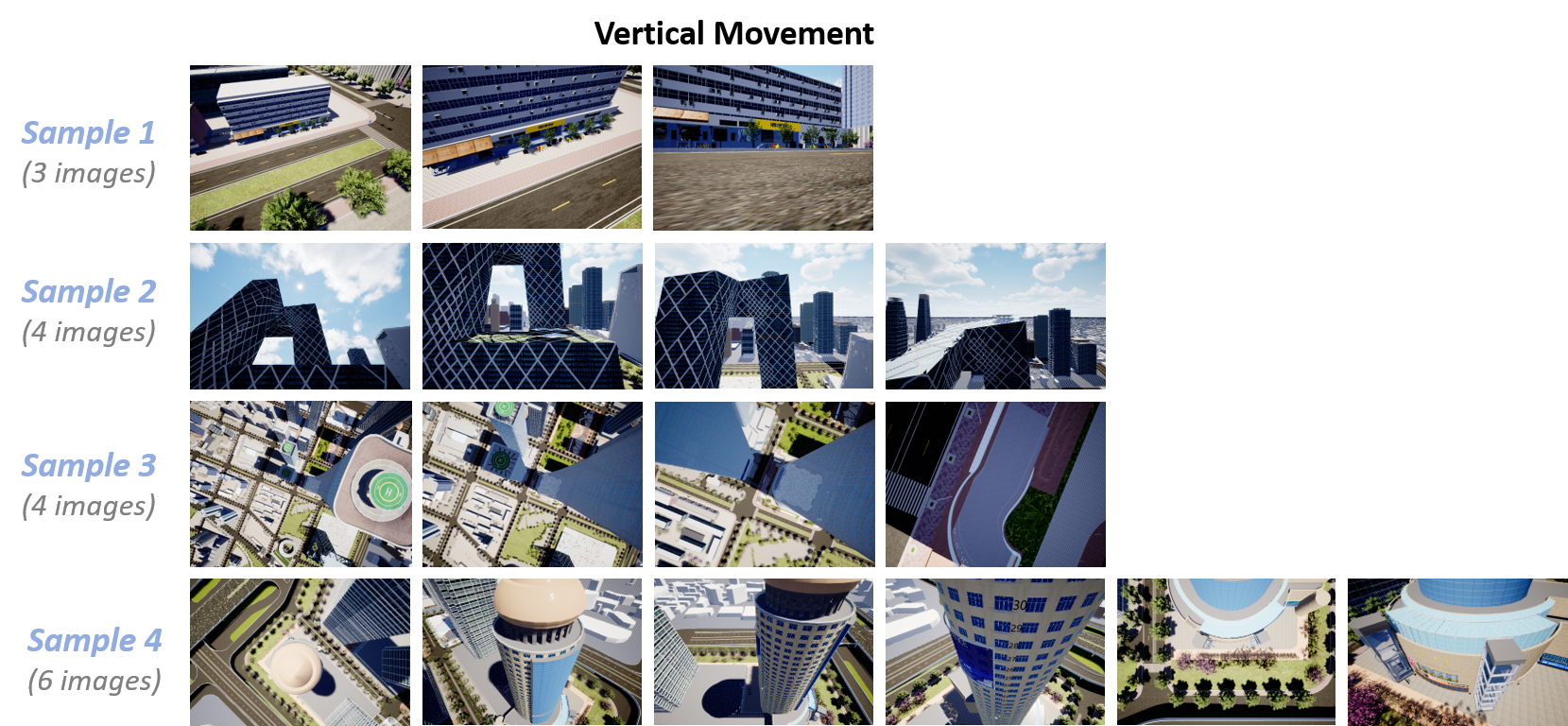}
        \caption{Multi-scale views in vertical movement during dynamic translation.}
    \end{subfigure}

    \caption{Examples of manually captured simulated images under dynamic translation with different motion patterns.}
    \label{fig:manual_examples}
\end{figure}

As shown in Fig.~\ref{fig:manual_examples}(a), we select a visually distinctive target as an anchor point, observe it from a long distance, gradually approach it, and finally localize the target for close-range observation. During this process, the number of captured images is not fixed (typically 4--8), but is instead determined by the observability of the target.

As illustrated in Fig.~\ref{fig:manual_examples}(b), we perform takeoff or landing from an open plaza with salient landmarks. Throughout this process, we record the changes in perceived urban scale, with the number of images ranging from 3 to 6.

\begin{figure}
\centering
\includegraphics[width=\linewidth]{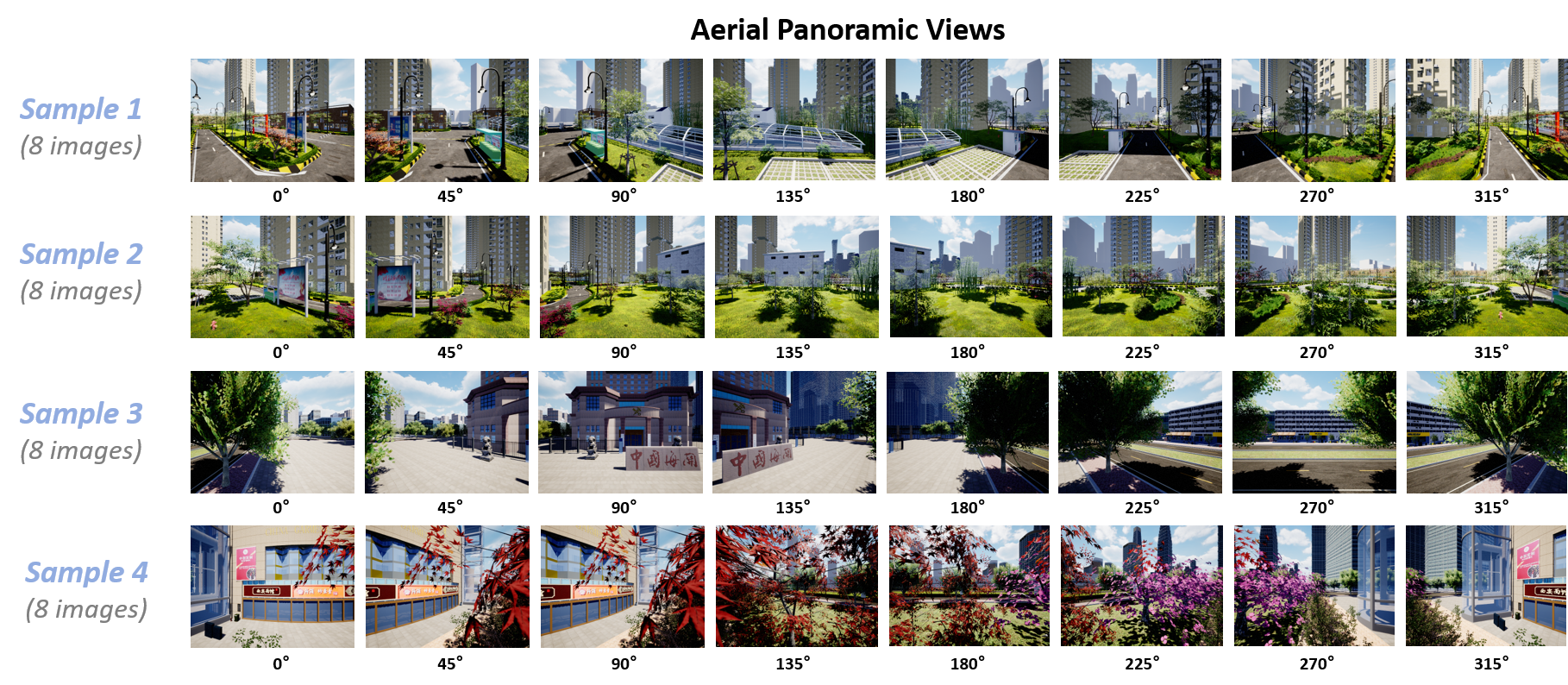}
\caption{Examples of simulated images captured manually in rotation (panoramic views).}
\label{fig:manual_example3}
\end{figure}

As shown in Fig.~\ref{fig:manual_example3}, we collected 42 sets of panoramic observations at multiple locations in EmbodiedCity. Each set consists of eight camera images with a 90-degree field of view, which together form a complete 360-degree observation for the aerial agent.

\subsection{Details of Multi-choice QA Generation}
\subsubsection{Role Templates}
\label{subsec:role_playing}
We adopt explicit role templates to frame the VLM as an embodied agent or an expert evaluator,
which has been shown to be critical for inducing structured reasoning behaviors and consistent
input--output formats in large language models. 

\begin{figure}
\centering
\includegraphics[width=\linewidth]{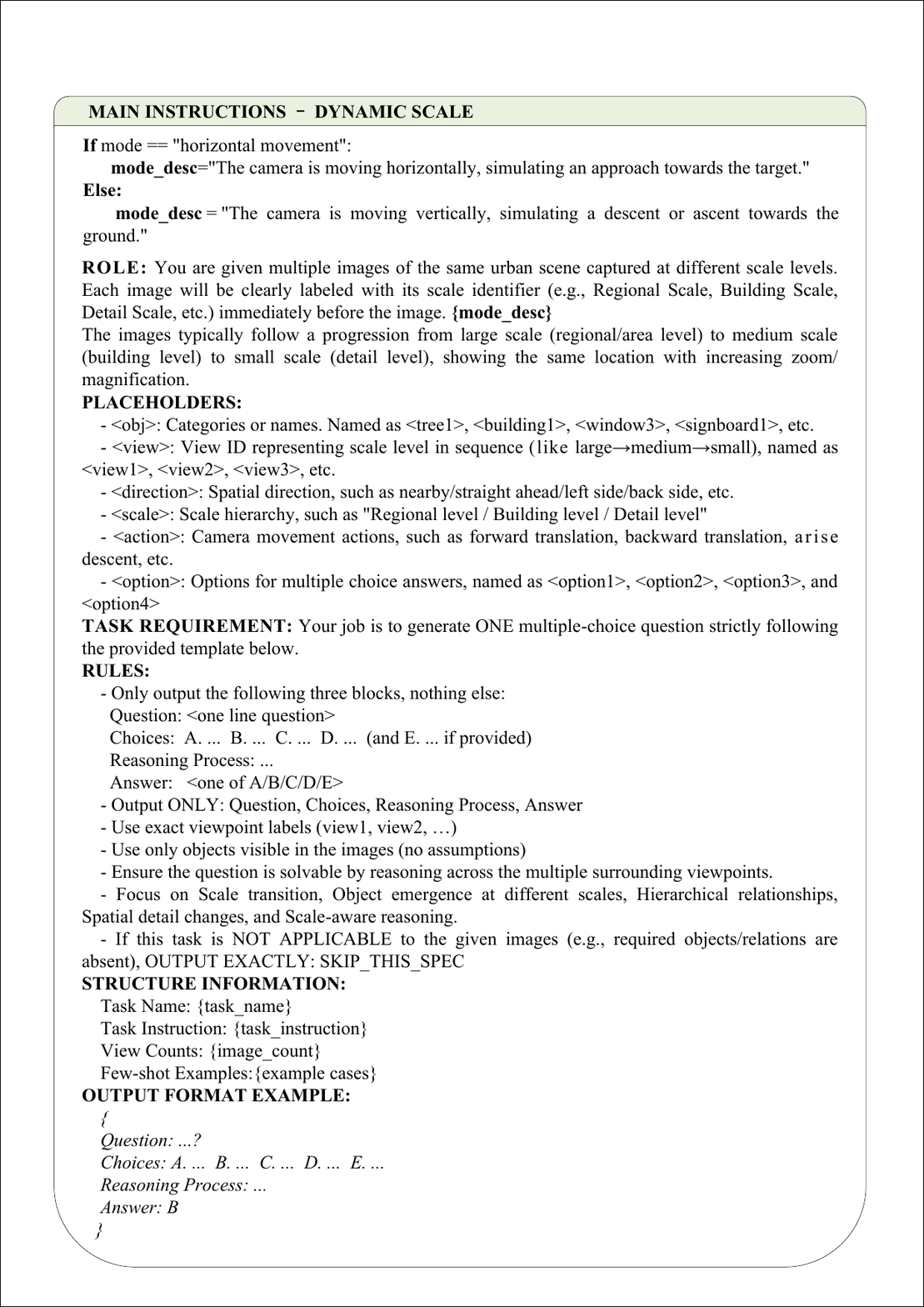}
\caption{Prompt template for role playing (Dynamic-scale views).}
\label{fig:main_prompt_ds}
\end{figure}

\textbf{Multi-scale Role (Fig.~\ref{fig:main_prompt_ds}).}
The first role template frames the model as an observer reasoning across a hierarchy of spatial
scales, ranging from regional to building and detail levels.
By explicitly labeling scale transitions and enforcing scale-aware placeholders, this template
encourages hierarchical spatial reasoning and object emergence analysis, which aligns with
human spatial cognition theories emphasizing multi-level environmental representation.

\begin{figure}
\centering
\includegraphics[width=\linewidth]{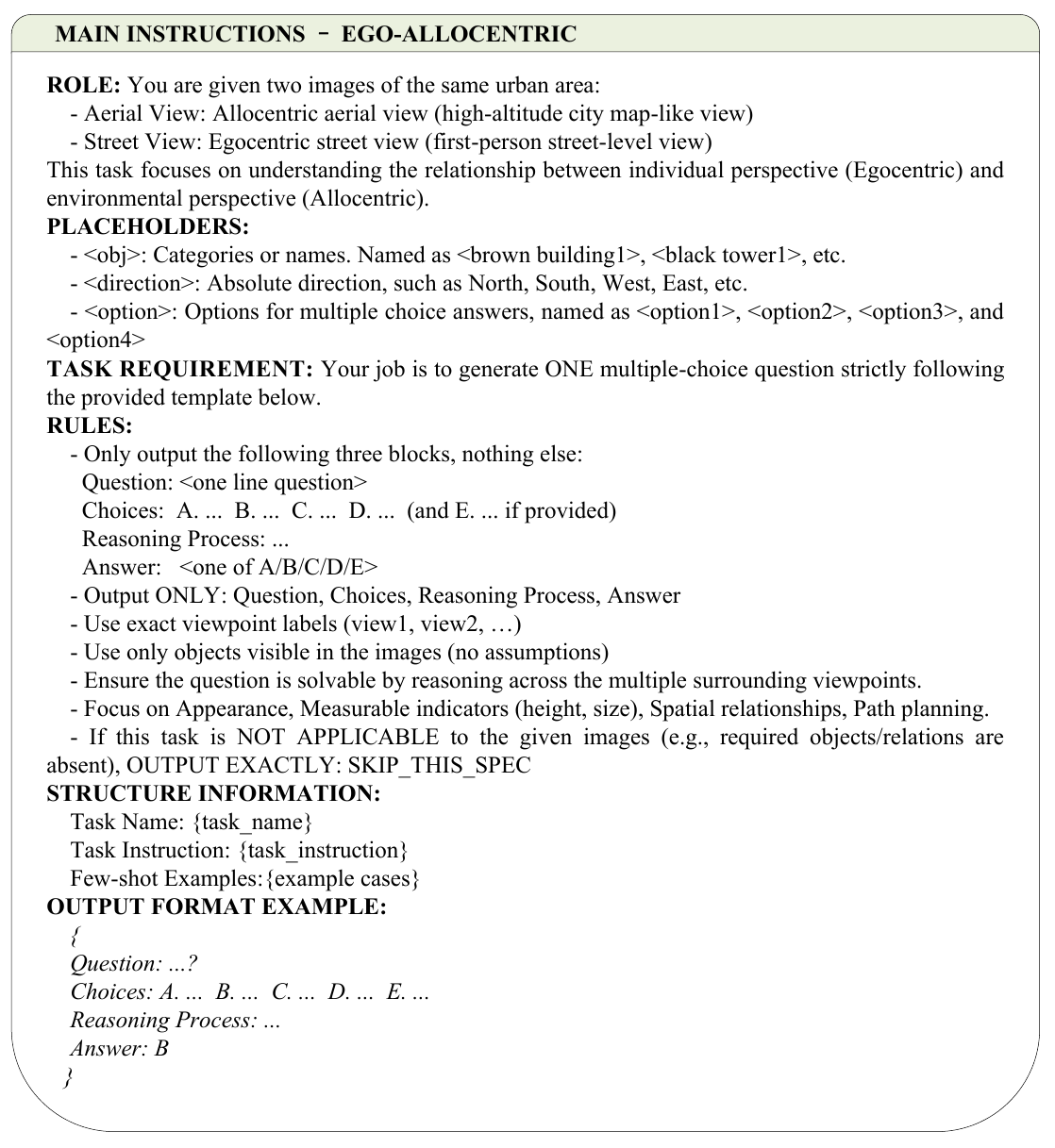}
\caption{Prompt template for role playing (Ego-allocentric views).}
\label{fig:main_prompt_egoallo}
\end{figure}

\textbf{Egocentric--Allocentric Role (Fig.~\ref{fig:main_prompt_egoallo}).}
The second role template assigns the model the task of jointly reasoning over egocentric street-level
views and allocentric aerial views of the same urban area.
This design explicitly bridges self-centered perception and map-like global understanding,
a distinction widely studied in cognitive science and embodied navigation.
Such a role formulation enables the model to align local observations with global spatial context.

\begin{figure}
\centering
\includegraphics[width=\linewidth]{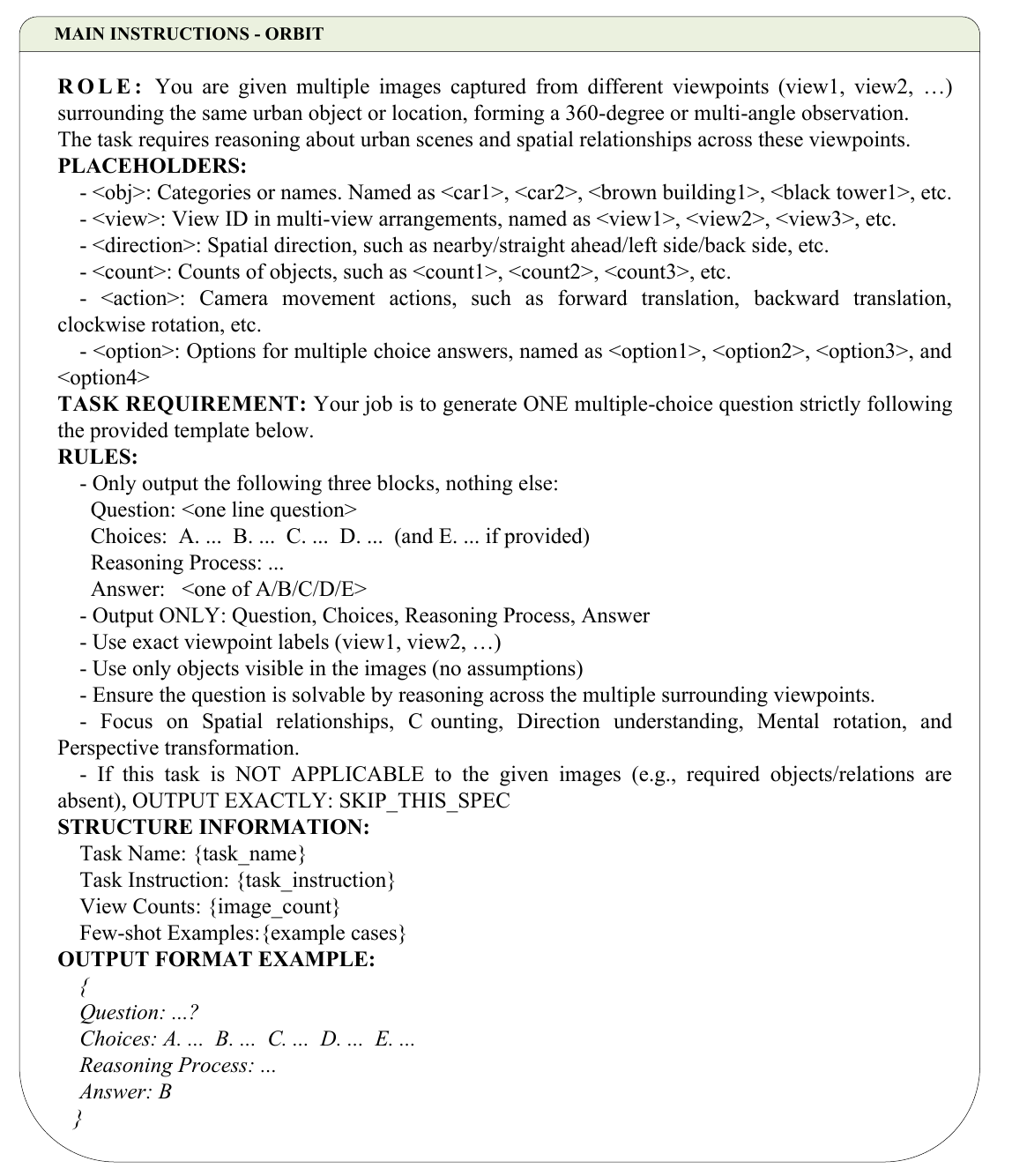}
\caption{Prompt template for role playing (Orbit views).}
\label{fig:main_prompt_orbit}
\end{figure}

\textbf{Surrounding Multi-view Role (Fig.~\ref{fig:main_prompt_orbit}).}
The third role template positions the VLM as an agent observing an urban object or location from
multiple surrounding viewpoints, forming a 360-degree or multi-angle observation.
This setting encourages cross-view consistency reasoning and spatial relation inference,
which is essential for robust scene understanding under viewpoint variations.

\begin{figure}
\centering
\includegraphics[width=\linewidth]{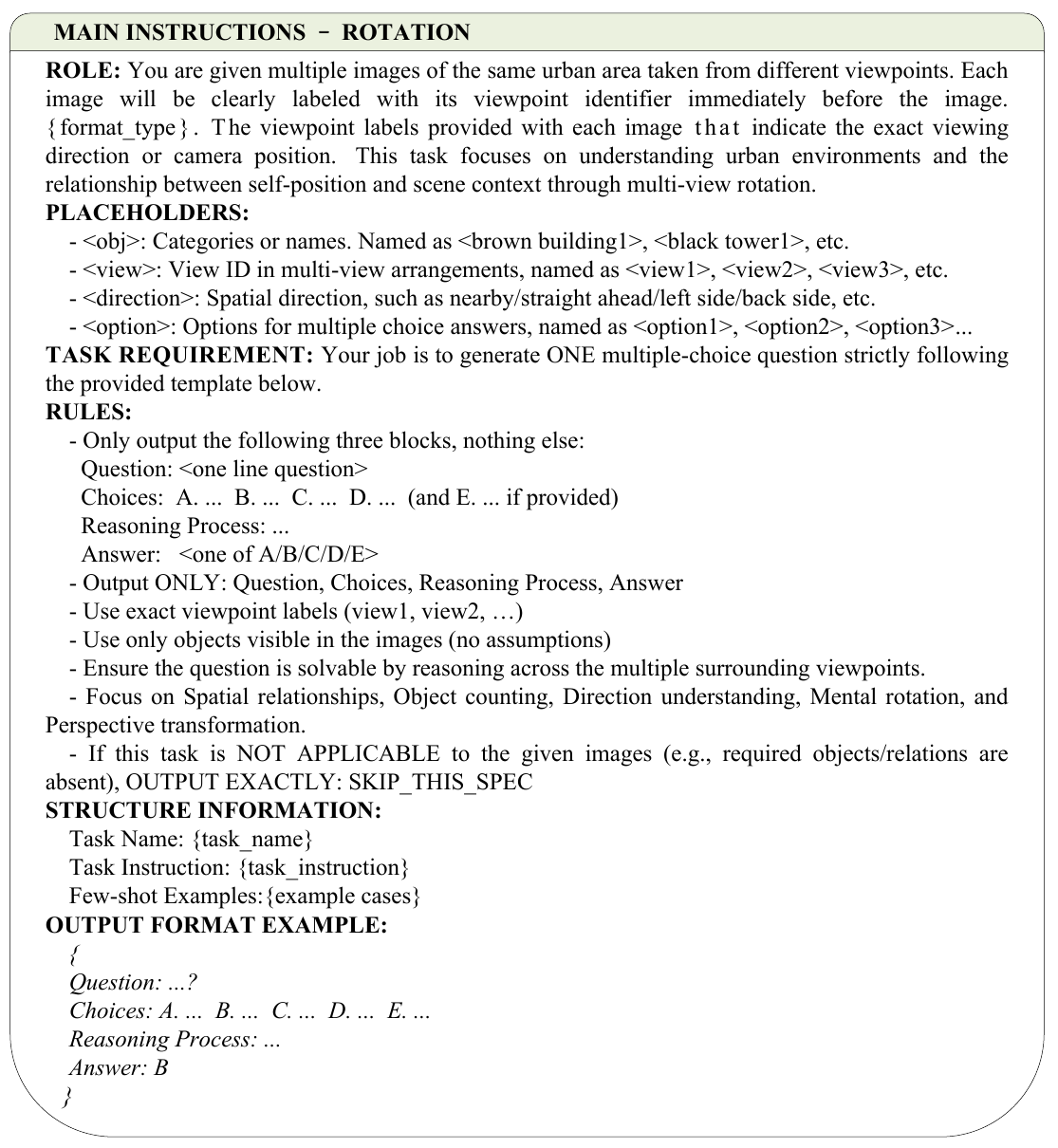}
\caption{Prompt template for role playing (Rotation views).}
\label{fig:main_prompt_rotation}
\end{figure}

\textbf{Rotation-based Viewpoint Role (Fig.~\ref{fig:main_prompt_rotation}).}
The fourth role template emphasizes rotational viewpoint changes around a fixed location,
with each image annotated by precise viewing directions or camera poses.
By enforcing strict viewpoint identifiers, this role promotes reasoning about self-position
changes relative to a static environment, supporting rotation-aware spatial inference and
mental viewpoint transformation.

\subsubsection{Task Templates}
\begin{figure*}
\centering
\includegraphics[width=\linewidth]{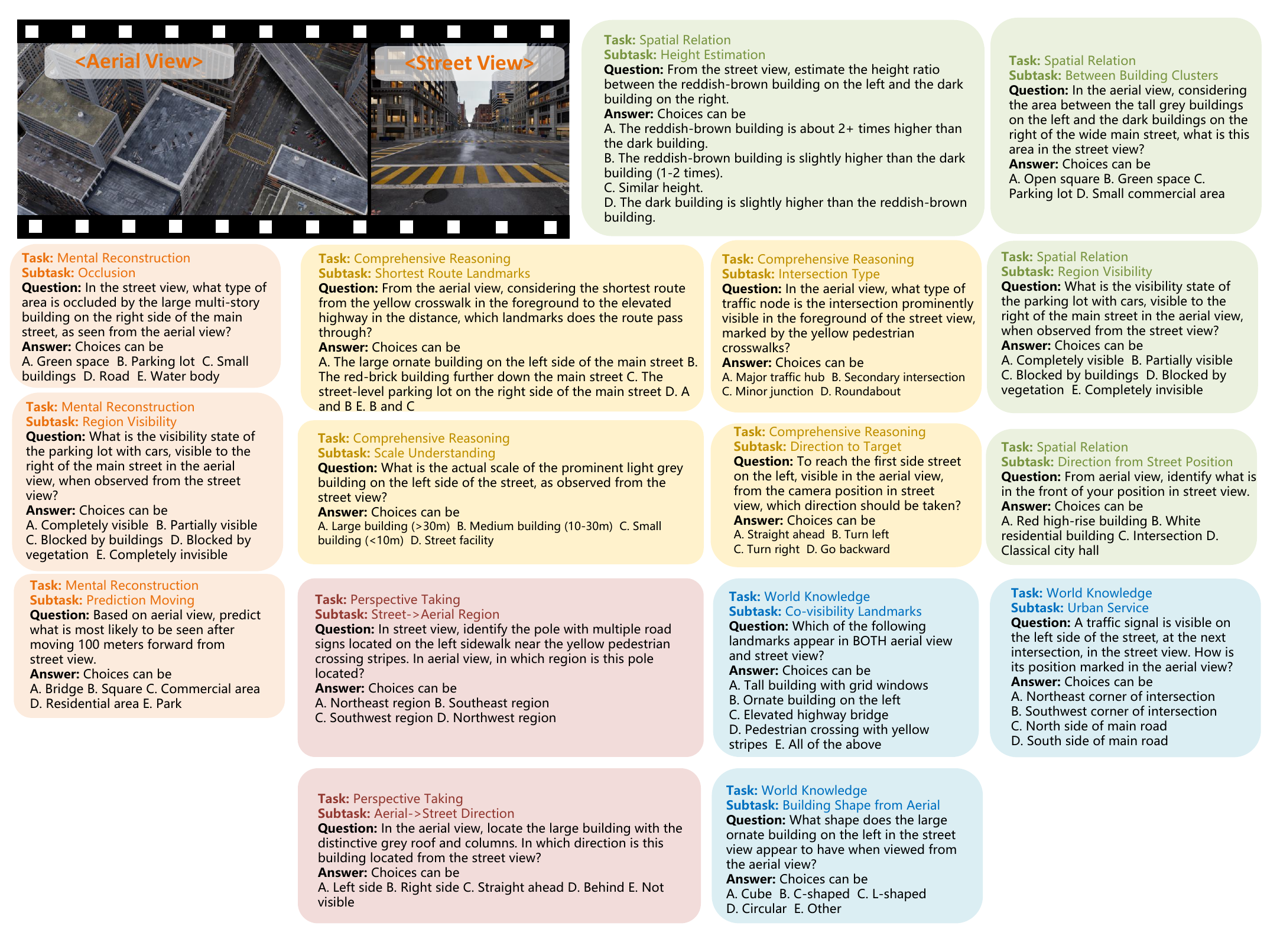}
\caption{Examples of task templates and their corresponding images (Ego-allocentric views).}
\label{egoallotem}
\end{figure*}

\begin{figure*}
\centering
\includegraphics[width=\linewidth]{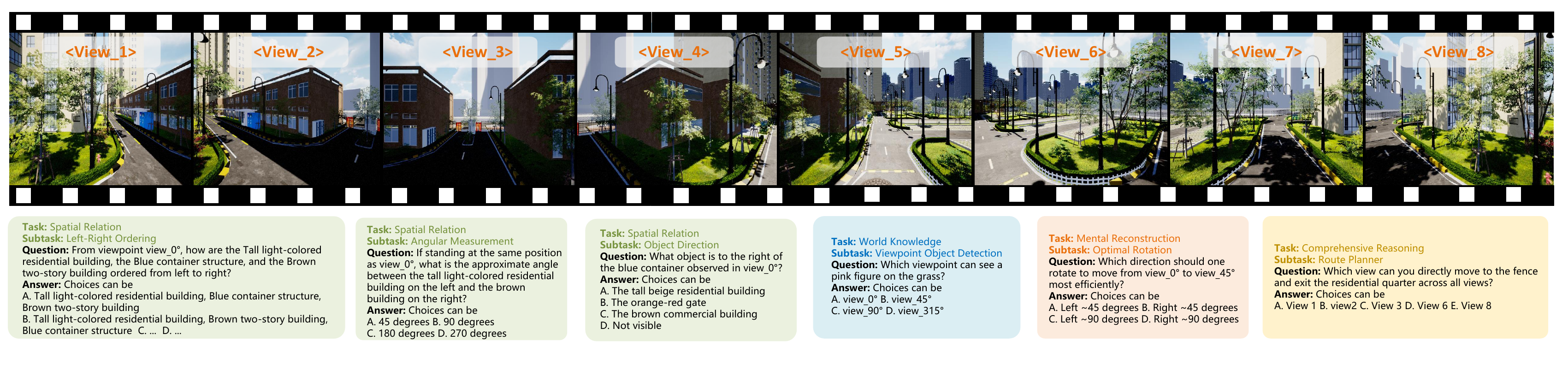}
\caption{Examples of simulated images captured manually in rotation (Rotation views).}
\label{rotationtem}
\end{figure*}

As shown in Fig \ref{egoallotem} to \ref{rotationtem}, we sequentially present the task templates for each camera dynamic, because these tasks are unique to this viewpoint mode.

\begin{figure*}
\centering
\includegraphics[width=\linewidth]{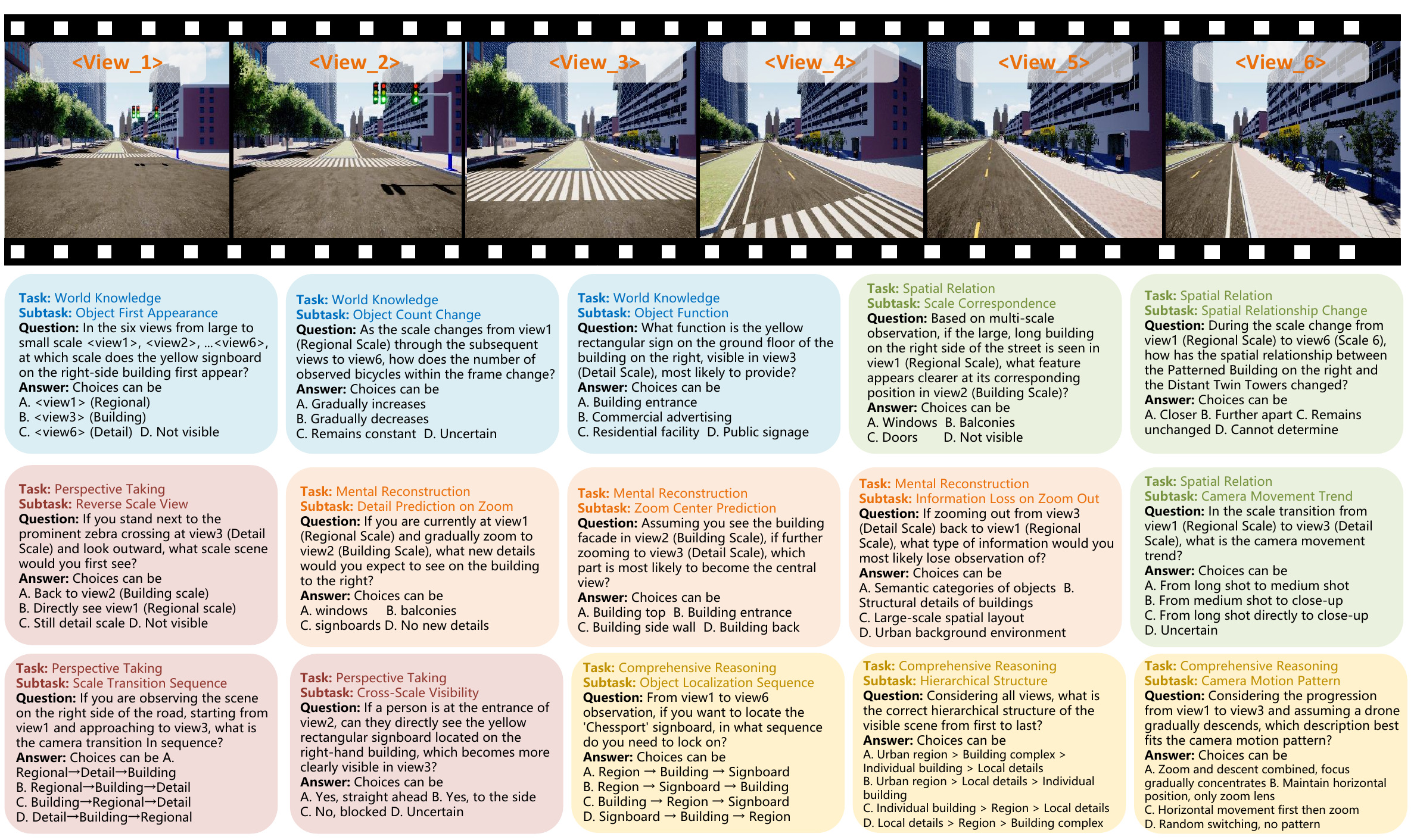}
\caption{Examples of simulated images captured manually in rotation (Orbit views).}
\end{figure*}

\begin{figure*}
\centering
\includegraphics[width=\linewidth]{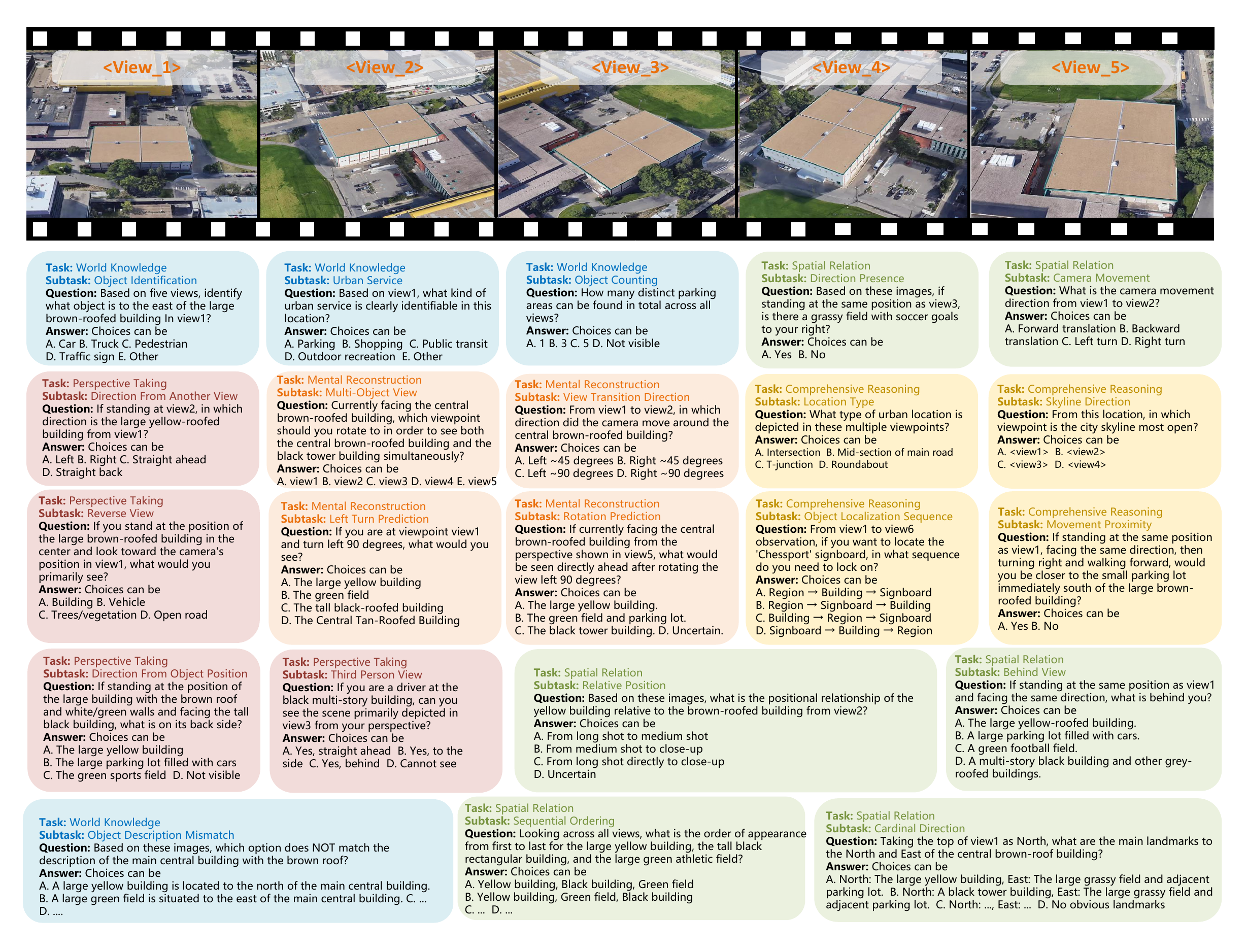}
\caption{Examples of simulated images captured manually in rotation (panoramic views).}
\end{figure*}

\subsection{Quality Control}

\subsubsection{Blind Filter} This filtering strategy aims to exclude questions that can be resolved solely through commonsense reasoning, without relying on explicit visual evidence. Specifically, we employ multiple VLMs (six open-source models in this study) to answer each question without providing any multi-view image inputs. If all VLMs correctly predict the answer under this setting, the question is discarded, indicating that it can be solved using general perceptual priors alone and does not require detailed visual interpretation of the scene.

This procedure ensures that the retained questions genuinely demand complex image-grounded reasoning and the integration of information across multiple viewpoints, thereby sharpening the dataset’s emphasis on challenging visual perception tasks. Conversely, questions for which all VLMs fail to produce the correct answer are also removed, as such cases suggest that the question may be ill-posed, ambiguous, or outside the intended scope of the task.

Together, this bidirectional filtering mechanism ensures that the final dataset consists exclusively of questions that require authentic multi-view visual reasoning—namely, those that cannot be trivially answered by all models, yet remain solvable by some VLMs when visual input is absent.
\subsubsection{Details of Human Refinement}
\label{sec:humanrefine}

This stage adopts a structured two-stage human refinement protocol to ensure annotation reliability and consistency. We recruit volunteers on campus and pay them reasonable compensation commensurate with the region. In detail, we recruit ten annotators with master's or doctoral training and research experience related to urban spatial understanding. Annotators are divided into two independent groups with complementary roles: a \emph{refinement group}, responsible for revising automatically generated QA pairs, and a \emph{verification group}, responsible for independent validation and consistency checking.

Each QA instance is reviewed by at least one annotator from each group, ensuring dual human coverage. Disagreements between the two groups are explicitly recorded and resolved through adjudication by the verification group, which serves as the final decision authority. This process implicitly enforces inter-annotator consistency by requiring agreement across independent reviewers before acceptance.

During refinement, annotators explicitly flag ambiguous or ill-posed questions, including cases with unclear spatial references, underspecified viewpoints, or multiple plausible answers. Such cases are either revised through rewording and answer option correction or removed entirely if ambiguity cannot be reliably resolved. This auditing process prevents semantically underspecified samples from entering the benchmark.

To further characterize refinement outcomes, we categorize common annotation issues encountered during this stage, including (i) incorrect spatial relations, (ii) misleading or visually unsupported answer options, (iii) inconsistent reasoning chains, and (iv) viewpoint-dependent ambiguities. Accepted QA pairs undergo final proofreading to eliminate residual ambiguities and formatting errors, and human-authored reasoning processes are added to improve clarity and interpretability.

\begin{figure*}
\centering
\includegraphics[width=\linewidth]{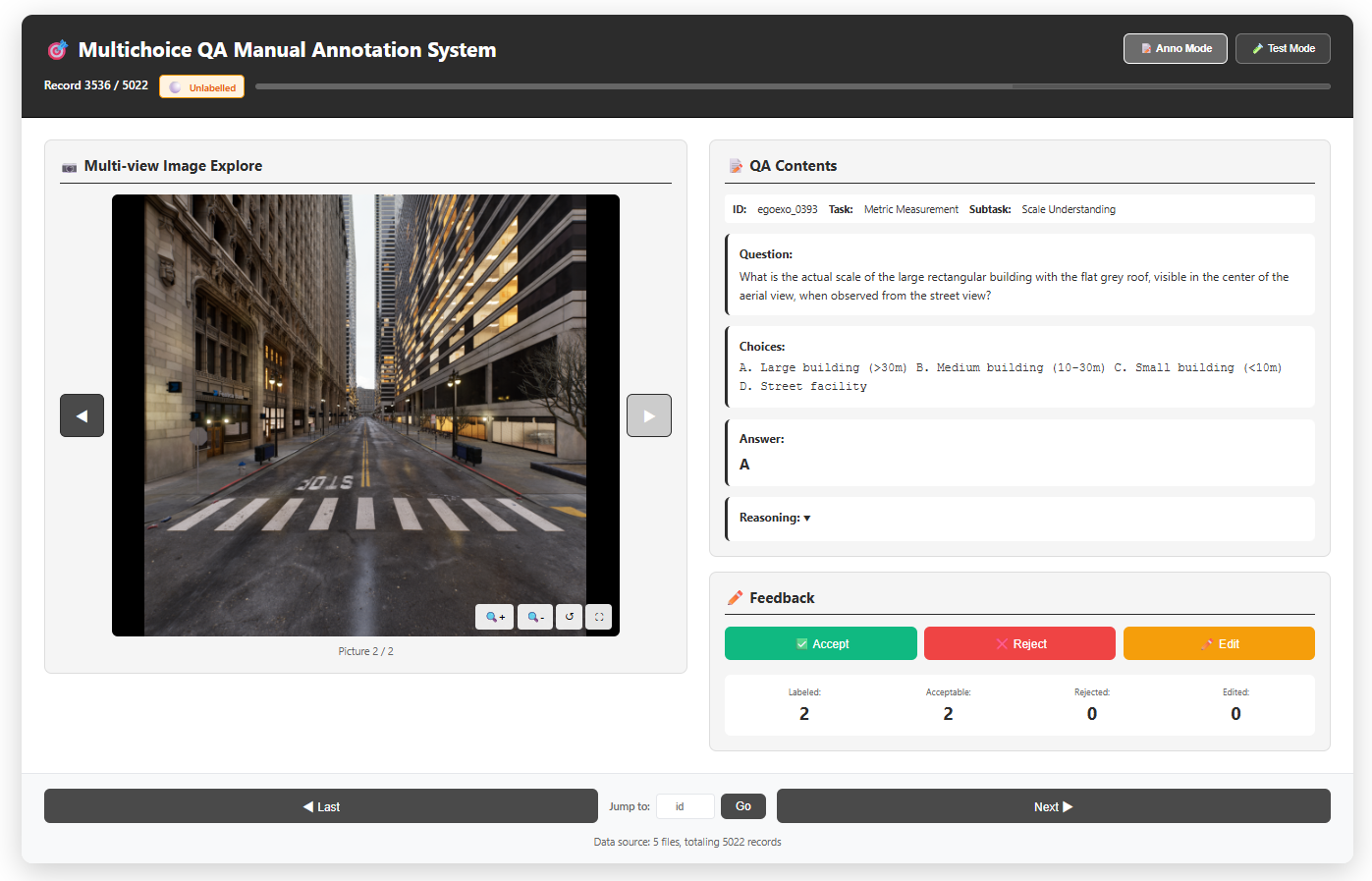}
\caption{A demonstration Website of Multi-choice QA Annotation System on CityCube Dataset.}
\label{fig:human_refine}
\end{figure*}

The refinement interface used by annotators is illustrated in Fig.~\ref{fig:human_refine}.

Overall, this refinement protocol prioritizes annotation reliability over scale, ensuring that subtle geometric relations are consistently grounded in the provided multi-view visual evidence.

%

\section{Experimental Details}
\label{sec:exp_details}
All local model inference and fine-tuning is performed on 4×A100-SXM4-80GB. The code of our program is available at \href{https://anonymous.4open.science/r/CityCube-Bench-9E72/}{anonymous.4open.science/r/CityCube-Bench-9E72/}.


\subsection{Brief Introduction on Baselines}
Our evaluation covers both proprietary and open-source
MLLMs trained to receive multi-image inputs. The evaluated
models, as well as random and human baselines, are briefly
introduced as follows:

\textbf{Random.}
A random baseline model serving as the lowest performance
benchmark for comparison.

\textbf{Human.}
Human expert performance baseline representing the upper
limit of human-level capability on this task.

\textbf{GPT-5.1.}
A state-of-the-art proprietary multimodal large language
model from OpenAI \cite{openai2025gpt5systemcard}, featuring advanced visual reasoning
and long-context understanding capabilities.

\textbf{Gemini-2.5-Pro.}
A high-performance multimodal model developed by Google \cite{comanici2025gemini},
designed for complex reasoning over visual and textual
inputs with strong generalization ability.

\textbf{Qwen-3-VL-Plus.}
A large-scale vision--language model from Alibaba \cite{bai2025qwen3vltechnicalreport}, optimized
for multi-image understanding and instruction-following
across diverse visual reasoning tasks.

\textbf{Step-1o-turbo-vision.}
A reasoning-optimized vision-language model from StepFun,
demonstrating competitive performance on multi-step and
compositional visual reasoning benchmarks.

\textbf{Doubao-seed1.6-251015.}
A proprietary multimodal model from ByteDance \cite{guo2025seed15vltechnicalreport}, designed
to support general-purpose vision--language understanding
and reasoning tasks.

\textbf{Skywork-R1V4-Lite.}
The largest vision--language reasoning model from SkyworkAI \cite{wei2023skywork},
focusing on high-performance inference while maintaining strong
visual comprehension capabilities.

\textbf{Qwen3-VL.}
An open-source vision--language model family supporting
multi-image inputs, widely adopted as a strong baseline
for multimodal reasoning and perception tasks.

\textbf{GLM-4.1V.}
A multimodal extension of the GLM series, featuring
enhanced visual understanding and cross-modal reasoning
abilities.

\textbf{Kimi-VL-A3B.}
A compact vision--language model emphasizing efficient
visual perception and instruction-following under limited
parameter budgets.

\textbf{MiMo-VL-7B.}
A 7B-parameter open-source vision--language model designed
for general multimodal understanding and reasoning.

\textbf{MiniCPM-V-4.5.}
A lightweight multimodal model from the MiniCPM series,
targeting efficient deployment with competitive visual
reasoning performance.

\textbf{Ovis2.5.}
A medium-scale vision--language model from Alibaba, supporting multi-image
inputs and fine-grained visual reasoning.

\textbf{LLaVA-NeXT-Video.}
An extension of LLaVA-NeXT tailored for video and multi-frame
visual reasoning, enabling temporal understanding across
sequential visual inputs.

\textbf{LLaVA-OneVision.}
A unified vision--language model supporting multiple visual
modalities and tasks within a single architecture.

\textbf{InternVL2.5.}
A strong open-source vision--language model with enhanced
multi-image reasoning and long-context modeling capabilities.

\textbf{Skywork-VL.}
A general-purpose open-source vision--language model designed
for multi-modal perception and reasoning tasks.

\textbf{Molmo-7B.}
A 7B-parameter multimodal model from AllenAI, focusing on robust visual
understanding and reasoning across diverse scenarios.

\textbf{Phi-4-multimodal-instruct.}
A compact multimodal instruction-tuned model from Microsoft, emphasizing
reasoning efficiency and controllability.

\textbf{Spatial-SSRL.}
A specialized vision--language model designed for spatial
reasoning and structured scene understanding.

\textbf{SpaceOm.}
A spatially-aware multimodal model focusing on object-level
and relational reasoning in complex visual environments.

\textbf{SpaceThinker.}
A reasoning-centric vision--language model explicitly
designed to enhance spatial cognition and multi-step
visual reasoning.

\section{Detailed Error Analysis}
\label{sec:error_details}
\begin{figure*}
\centering
\includegraphics[width=.95\linewidth]{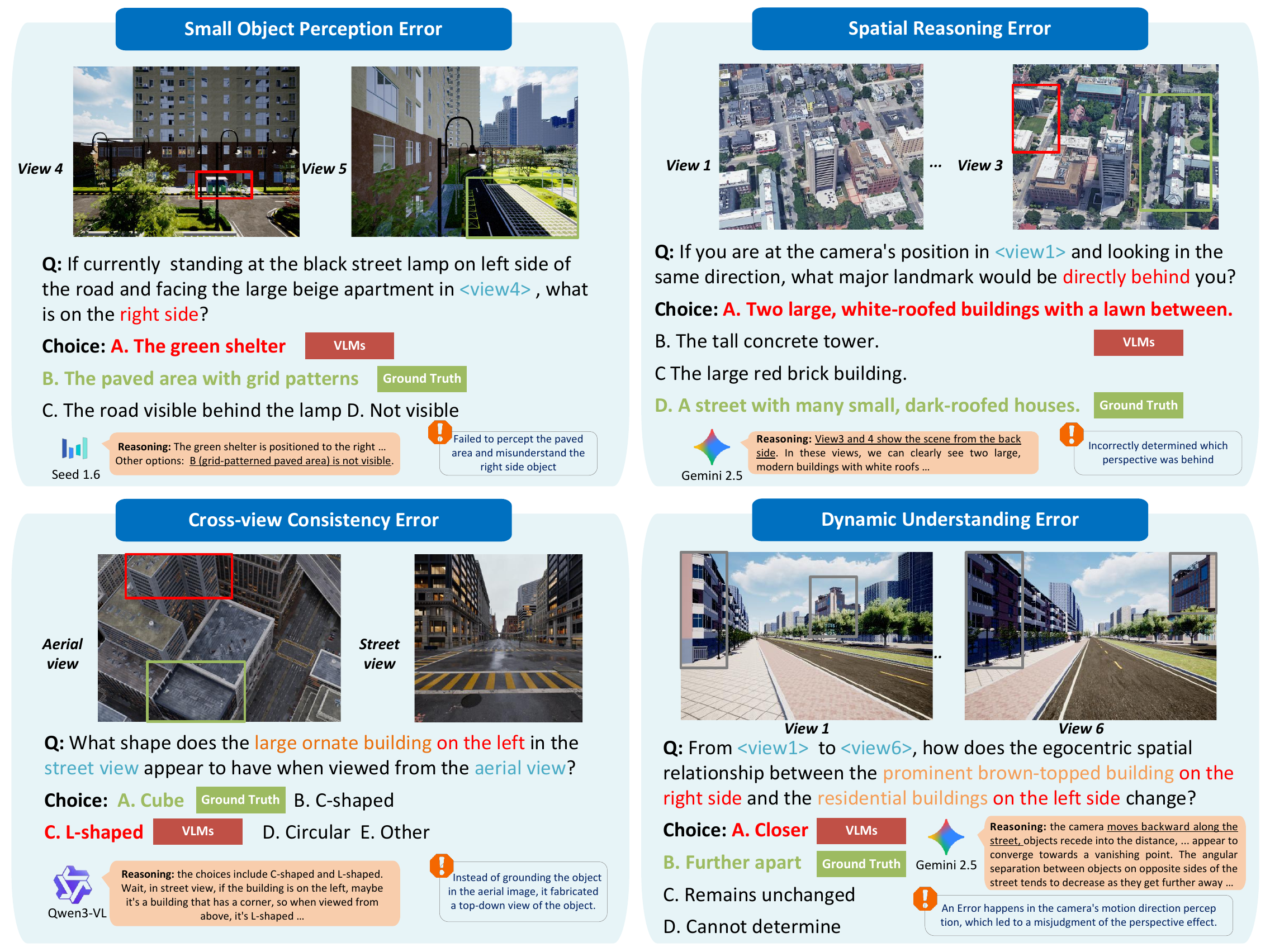}
\caption{The four common failure cases for VLMs in multi-view spatial reasoning.}
\label{fig:error_analysis}
\end{figure*}
To provide a deeper understanding of the challenges faced by VLMs in urban spatial reasoning, we detail the specific failure cases illustrated in Fig. \ref{fig:error_analysis}. These examples correspond to the four primary error modes discussed in the main text.

\subsection{Small Object Perception Error} Case Description: As shown in the first column of Fig. \ref{fig:error_analysis}, the model (Seed 1.6) was tasked with identifying the object located on the right side of a specific viewpoint in a street scene (View 4). The ground truth answer is "The paved area with grid patterns" visible in View 5. Failure Analysis: The model incorrectly selected "The green shelter" , explicitly stating in its reasoning that the grid-patterned paved area "is not visible". This demonstrates a failure to perceive low-salience or small-scale objects (the paved area) within the complex urban scene, leading to a hallucinated spatial arrangement where it defaulted to a more prominent object (the shelter) despite it not being the correct answer for the specific spatial query.

\subsection{Spatial Reasoning Error} Case Description: In the second case, the model (Gemini 2.5 pro) was asked to determine what major landmark lies directly behind the camera's position in an aerial shot (View 1). The correct answer, based on the global context, is "A street with many small, dark-roofed houses". Failure Analysis: The model incorrectly chose "Two large, white-roofed buildings". The reasoning reveals that the model attempted to deduce the rear view by referencing other views (View 3 and 4) , but it failed to correctly orient these views relative to View 1. It stated that these views show the scene from the "back side," which was a misinterpretation of the spatial layout. This highlights the model's inability to perform accurate mental rotation or establish a coherent 360-degree spatial representation from disjointed views.

\subsection{Cross-view Consistency Error} Case Description: The third column illustrates a task requiring the model (Qwen3-VL) to determine the shape of a building seen in a street view when viewed from above (Aerial view). The target building is a "Cube". Failure Analysis: The model incorrectly reasoned that the building is "L-shaped". The reasoning log shows that instead of visually grounding the specific building in the provided aerial image, the model relied on a heuristic from the street view perspective ("maybe it's a building that has a corner"). This led to a fabrication of the top-down geometry rather than a retrieval of visual evidence from the aerial modality, exemplifying a disconnection between the ground-level and overhead perspectives.

\subsection{Dynamic Understanding Error} Case Description: In the final example involving motion dynamics, the model (Gemini 2.5 pro) was asked to judge how the spatial relationship (distance) between buildings on opposite sides of the street changes as the camera moves from View 1 to View 6. The ground truth is that they appear "Further apart" due to the perspective change. Failure Analysis: The model incorrectly predicted the objects would get "Closer". The reasoning indicates a fundamental error in perceiving the direction of motion; the model believed the camera was moving "backward" and that objects were receding. In reality, the camera motion or perspective shift was misinterpreted, leading to a reversed understanding of the perspective effect.

\section{Implementation Details}
\label{sec:humaneval}
The dataset is split into training and test sets with a ratio of 9:1, and data loading is
parallelized using four worker processes. The fine-tune experiment results are produced on test set.
\begin{table}[t]
\centering
\caption{Training hyperparameters for \textbf{CityBot} series.}
\label{tab:qwen3vl_hyperparams}
\resizebox{\linewidth}{!}{
\begin{tabular}{l l l}
\hline
\textbf{Category} & \textbf{Hyperparameter} & \textbf{Value} \\
\hline
\multirow{3}{*}{Training Setup}
 & Training paradigm & Supervised fine-tuning (LoRA) \\
 & Number of epochs & 5 \\
 & Precision & bfloat16 \\
\hline
\multirow{4}{*}{Batching}
 & Per-device train batch size & 1 \\
 & Per-device eval batch size & 1 \\
 & Gradient accumulation steps & 4 \\
 & Effective batch size & 4 \\
\hline
\multirow{2}{*}{Optimization}
 & Learning rate & $1\times10^{-4}$ \\
 & Warmup ratio & 0.05 \\
\hline
\multirow{4}{*}{LoRA Configuration}
 & LoRA rank ($r$) & 8 \\
 & LoRA scaling factor ($\alpha$) & 32 \\
 & Target modules & All linear layers \\
 & Training scope & Full model (no freezing) \\
\hline
\multirow{4}{*}{Architecture \& Memory}
 & Attention implementation & FlashAttention \\
 & Padding-free training & Enabled \\
 & Sample packing & Enabled \\
 & Gradient checkpointing & Enabled (ViT excluded) \\
\hline
\multirow{2}{*}{Input}
 & Maximum sequence length & 4096 \\
 & Maximum image pixels & 1,605,632 \\
\hline
\multirow{4}{*}{Evaluation \& Logging}
 & Evaluation interval & Every 100 steps \\
 & Checkpoint saving interval & Every 100 steps \\
 & Max checkpoints kept & 3 \\
 & Logging interval & Every 10 steps \\
\hline
\multirow{3}{*}{Data Loading}
 & Test split ratio & 0.1 \\
 & Dataset preprocessing workers & 4 \\
 & Dataloader workers & 4 \\
\hline
\end{tabular}
}
\end{table}

\subsection{Training Details}
We fine-tune \textbf{Qwen3-VL-2B}, \textbf{Qwen3-VL-4B} and \textbf{Qwen3-VL-8B} using supervised fine-tuning with Low-Rank Adaptation (LoRA).
As described in Tab \ref{tab:qwen3vl_hyperparams}, the models are trained for 5 epochs using the Adam optimizer with an initial learning rate
of $1\times10^{-4}$ and a warmup ratio of 0.05.
To reduce memory consumption while maintaining training stability, we employ bfloat16 precision,
gradient accumulation with 4 steps, and FlashAttention.
The maximum input sequence length is set to 4096 tokens, and images are resized to ensure the total
pixel count does not exceed 1.6M.

LoRA is applied to all linear layers with a rank of 8 and a scaling factor of 32.
We do not freeze the vision encoder or the
vision-language alignment module to allow full end-to-end adaptation.
We further enable padding-free training and sample packing to improve computational efficiency.

During training, evaluation and checkpoint saving are performed every 100 steps, and at most
three checkpoints are retained.

\subsection{Human Evaluation Details}
\begin{figure*}
\centering
\includegraphics[width=\linewidth]{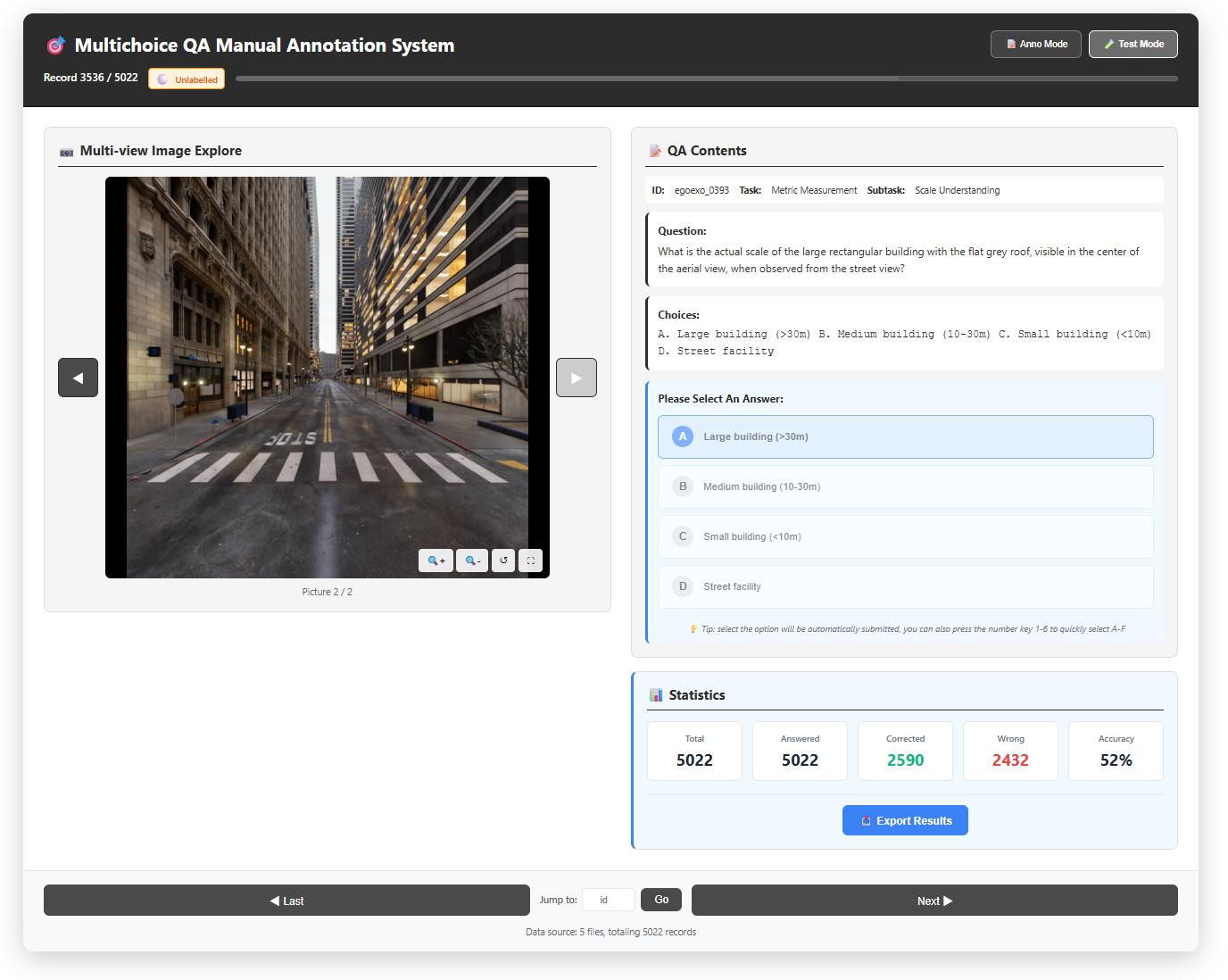}
\caption{A demonstration Website of Multi-choice QA Test System of CityCube Dataset.}
\label{fig:human_eval}
\end{figure*}

As shown in Figure \ref{fig:human_eval}, we collected a set of volunteer evaluation results through an interactive web interface. These volunteers are recruited independently and unrelated to the annotators who participated in refining the questions. They also did not receive any extra training in spatial knowledge.

\section{Further Discussion}
\label{sec:furtherdiscuss}
\textbf{3D Question Answering}
A precise understanding of objects and their spatial relationships within 3D scenes is critical for applications in robotics and autonomous driving. Leveraging the holistic 3D scene datasets such as ScanNet \cite{dai2017scannet} and Matterport3D \cite{chang2017matterport3d}, the research community has developed a variety of 3D Question Answering (3DQA) benchmarks like ScanQA \cite{azuma2022scanqa, achlioptas2020referit3d, hong20233d}. The core paradigm of these benchmarks involves inferring a comprehensive 3D understanding from a limited set of 2D views.
In line with this, employing multiple views as input for the LLMs has become a primary methodology in 3DQA \cite{fu2024scene, huang2024embodied, guo2023viewrefer}. Our work extends this research trajectory and systematically incorporates typical view combinations found in existing literature and, more importantly, pioneers the extension of the 3DQA application domain from indoor environments to the more expansive and challenging context of urban spaces.

%% file: custom.bib
@inproceedings{majumdar2024openeqa,
  title={Openeqa: Embodied question answering in the era of foundation models},
  author={Majumdar, Arjun and Ajay, Anurag and Zhang, Xiaohan and Putta, Pranav and Yenamandra, Sriram and Henaff, Mikael and Silwal, Sneha and Mcvay, Paul and Maksymets, Oleksandr and Arnaud, Sergio and others},
  booktitle={Proceedings of the IEEE/CVF conference on computer vision and pattern recognition},
  pages={16488--16498},
  year={2024}
}

@inproceedings{yang2024depth,
  title={Depth anything: Unleashing the power of large-scale unlabeled data},
  author={Yang, Lihe and Kang, Bingyi and Huang, Zilong and Xu, Xiaogang and Feng, Jiashi and Zhao, Hengshuang},
  booktitle={Proceedings of the IEEE/CVF conference on computer vision and pattern recognition},
  pages={10371--10381},
  year={2024}
}

@inproceedings{cai2025spatialbot,
  title={Spatialbot: Precise spatial understanding with vision language models},
  author={Cai, Wenxiao and Ponomarenko, Iaroslav and Yuan, Jianhao and Li, Xiaoqi and Yang, Wankou and Dong, Hao and Zhao, Bo},
  booktitle={2025 IEEE International Conference on Robotics and Automation (ICRA)},
  pages={9490--9498},
  year={2025},
  organization={IEEE}
}

@inproceedings{chen2024spatialvlm,
  title={Spatialvlm: Endowing vision-language models with spatial reasoning capabilities},
  author={Chen, Boyuan and Xu, Zhuo and Kirmani, Sean and Ichter, Brain and Sadigh, Dorsa and Guibas, Leonidas and Xia, Fei},
  booktitle={Proceedings of the IEEE/CVF Conference on Computer Vision and Pattern Recognition},
  pages={14455--14465},
  year={2024}
}

@inproceedings{song2025robospatial,
  title={Robospatial: Teaching spatial understanding to 2d and 3d vision-language models for robotics},
  author={Song, Chan Hee and Blukis, Valts and Tremblay, Jonathan and Tyree, Stephen and Su, Yu and Birchfield, Stan},
  booktitle={Proceedings of the Computer Vision and Pattern Recognition Conference},
  pages={15768--15780},
  year={2025}
}

@book{tversky2019mind,
  title={Mind in motion: How action shapes thought},
  author={Tversky, Barbara},
  year={2019},
  publisher={Basic Books}
}

@article{ding2025understanding,
  title={Understanding world or predicting future? a comprehensive survey of world models},
  author={Ding, Jingtao and Zhang, Yunke and Shang, Yu and Zhang, Yuheng and Zong, Zefang and Feng, Jie and Yuan, Yuan and Su, Hongyuan and Li, Nian and Sukiennik, Nicholas and others},
  journal={ACM Computing Surveys},
  volume={58},
  number={3},
  pages={1--38},
  year={2025},
  publisher={ACM New York, NY}
}

@article{li2025viewspatial,
  title={ViewSpatial-Bench: Evaluating Multi-perspective Spatial Localization in Vision-Language Models},
  author={Li, Dingming and Li, Hongxing and Wang, Zixuan and Yan, Yuchen and Zhang, Hang and Chen, Siqi and Hou, Guiyang and Jiang, Shengpei and Zhang, Wenqi and Shen, Yongliang and others},
  journal={arXiv preprint arXiv:2505.21500},
  year={2025}
}

@book{piaget2013child,
  title={Child's conception of space: Selected works vol 4},
  author={Piaget, Jean},
  year={2013},
  publisher={Routledge}
}

@article{eslami2018neural,
  title={Neural scene representation and rendering},
  author={Eslami, SM Ali and Jimenez Rezende, Danilo and Besse, Frederic and Viola, Fabio and Morcos, Ari S and Garnelo, Marta and Ruderman, Avraham and Rusu, Andrei A and Danihelka, Ivo and Gregor, Karol and others},
  journal={Science},
  volume={360},
  number={6394},
  pages={1204--1210},
  year={2018},
  publisher={American Association for the Advancement of Science}
}

@article{jia2025omnispatial,
  title={OmniSpatial: Towards Comprehensive Spatial Reasoning Benchmark for Vision Language Models},
  author={Jia, Mengdi and Qi, Zekun and Zhang, Shaochen and Zhang, Wenyao and Yu, Xinqiang and He, Jiawei and Wang, He and Yi, Li},
  journal={arXiv preprint arXiv:2506.03135},
  year={2025}
}

@article{hong20233d,
  title={3d-llm: Injecting the 3d world into large language models},
  author={Hong, Yining and Zhen, Haoyu and Chen, Peihao and Zheng, Shuhong and Du, Yilun and Chen, Zhenfang and Gan, Chuang},
  journal={Advances in Neural Information Processing Systems},
  volume={36},
  pages={20482--20494},
  year={2023}
}

@inproceedings{zhang2024agent3d,
  title={Agent3d-zero: An agent for zero-shot 3d understanding},
  author={Zhang, Sha and Huang, Di and Deng, Jiajun and Tang, Shixiang and Ouyang, Wanli and He, Tong and Zhang, Yanyong},
  booktitle={European Conference on Computer Vision},
  pages={186--202},
  year={2024},
  organization={Springer}
}

@article{zhu2024llava,
  title={Llava-3d: A simple yet effective pathway to empowering lmms with 3d-awareness},
  author={Zhu, Chenming and Wang, Tai and Zhang, Wenwei and Pang, Jiangmiao and Liu, Xihui},
  journal={arXiv preprint arXiv:2409.18125},
  year={2024}
}

@inproceedings{qi2024shapellm,
  title={Shapellm: Universal 3d object understanding for embodied interaction},
  author={Qi, Zekun and Dong, Runpei and Zhang, Shaochen and Geng, Haoran and Han, Chunrui and Ge, Zheng and Yi, Li and Ma, Kaisheng},
  booktitle={European Conference on Computer Vision},
  pages={214--238},
  year={2024},
  organization={Springer}
}

@inproceedings{yang2025thinking,
  title={Thinking in space: How multimodal large language models see, remember, and recall spaces},
  author={Yang, Jihan and Yang, Shusheng and Gupta, Anjali W and Han, Rilyn and Fei-Fei, Li and Xie, Saining},
  booktitle={Proceedings of the Computer Vision and Pattern Recognition Conference},
  pages={10632--10643},
  year={2025}
}

@inproceedings{du2024embspatial,
  title={Embspatial-bench: Benchmarking spatial understanding for embodied tasks with large vision-language models},
  author={Du, Mengfei and Wu, Binhao and Li, Zejun and Huang, Xuan-Jing and Wei, Zhongyu},
  booktitle={Proceedings of the 62nd Annual Meeting of the Association for Computational Linguistics (Volume 2: Short Papers)},
  pages={346--355},
  year={2024}
}

@article{cheng2024spatialrgpt,
  title={Spatialrgpt: Grounded spatial reasoning in vision-language models},
  author={Cheng, An-Chieh and Yin, Hongxu and Fu, Yang and Guo, Qiushan and Yang, Ruihan and Kautz, Jan and Wang, Xiaolong and Liu, Sifei},
  journal={Advances in Neural Information Processing Systems},
  volume={37},
  pages={135062--135093},
  year={2024}
}

@article{zhao2025urbanvideo,
  title={Urbanvideo-bench: Benchmarking vision-language models on embodied intelligence with video data in urban spaces},
  author={Zhao, Baining and Fang, Jianjie and Dai, Zichao and Wang, Ziyou and Zha, Jirong and Zhang, Weichen and Gao, Chen and Wang, Yue and Cui, Jinqiang and Chen, Xinlei and others},
  journal={arXiv preprint arXiv:2503.06157},
  year={2025}
}

@article{zhang2025open3dvqa,
  title={Open3dvqa: A benchmark for comprehensive spatial reasoning with multimodal large language model in open space},
  author={Zhang, Weichen and Zhou, Zile and Zheng, Zhiheng and Gao, Chen and Cui, Jinqiang and Li, Yong and Chen, Xinlei and Zhang, Xiao-Ping},
  journal={arXiv preprint arXiv:2503.11094},
  year={2025}
}

@inproceedings{feng2025citygpt,
  title={Citygpt: Empowering urban spatial cognition of large language models},
  author={Feng, Jie and Liu, Tianhui and Du, Yuwei and Guo, Siqi and Lin, Yuming and Li, Yong},
  booktitle={Proceedings of the 31st ACM SIGKDD Conference on Knowledge Discovery and Data Mining V. 2},
  pages={591--602},
  year={2025}
}

@inproceedings{zhou2025urbench,
  title={Urbench: A comprehensive benchmark for evaluating large multimodal models in multi-view urban scenarios},
  author={Zhou, Baichuan and Yang, Haote and Chen, Dairong and Ye, Junyan and Bai, Tianyi and Yu, Jinhua and Zhang, Songyang and Lin, Dahua and He, Conghui and Li, Weijia},
  booktitle={Proceedings of the AAAI Conference on Artificial Intelligence},
  volume={39},
  pages={10707--10715},
  year={2025}
}

@article{feng2024citybench,
  title={Citybench: Evaluating the capabilities of large language model as world model},
  author={Feng, Jie and Zhang, Jun and Yan, Junbo and Zhang, Xin and Ouyang, Tianjian and Liu, Tianhui and Du, Yuwei and Guo, Siqi and Li, Yong},
  journal={arXiv e-prints},
  pages={arXiv--2406},
  year={2024}
}

@article{he2025urbanfeel,
  title={Urbanfeel: A comprehensive benchmark for temporal and perceptual understanding of city scenes through human perspective},
  author={He, Jun and Lin, Yi and Huang, Zilong and Yin, Jiacong and Ye, Junyan and Zhou, Yuchuan and Li, Weijia and Zhang, Xiang},
  journal={arXiv preprint arXiv:2509.22228},
  year={2025}
}

@article{liu2025citylens,
  title={CityLens: Benchmarking Large Language-Vision Models for Urban Socioeconomic Sensing},
  author={Liu, Tianhui and Feng, Jie and Pang, Hetian and Zhang, Xin and Ouyang, Tianjian and Zhang, Zhiyuan and Li, Yong},
  journal={arXiv preprint arXiv:2506.00530},
  year={2025}
}

@article{hao2024urbanvlp,
  title={Urbanvlp: A multi-granularity vision-language pre-trained foundation model for urban indicator prediction},
  author={Hao, Xixuan and Chen, Wei and Yan, Yibo and Zhong, Siru and Wang, Kun and Wen, Qingsong and Liang, Yuxuan},
  journal={CoRR},
  year={2024}
}

@article{feng2025urbanllava,
  title={UrbanLLaVA: A Multi-modal Large Language Model for Urban Intelligence with Spatial Reasoning and Understanding},
  author={Feng, Jie and Wang, Shengyuan and Liu, Tianhui and Xi, Yanxin and Li, Yong},
  journal={arXiv preprint arXiv:2506.23219},
  year={2025}
}

@inproceedings{dai2017scannet,
  title={Scannet: Richly-annotated 3d reconstructions of indoor scenes},
  author={Dai, Angela and Chang, Angel X and Savva, Manolis and Halber, Maciej and Funkhouser, Thomas and Nie{\ss}ner, Matthias},
  booktitle={Proceedings of the IEEE conference on computer vision and pattern recognition},
  pages={5828--5839},
  year={2017}
}

@article{chang2017matterport3d,
  title={Matterport3d: Learning from rgb-d data in indoor environments},
  author={Chang, Angel and Dai, Angela and Funkhouser, Thomas and Halber, Maciej and Niessner, Matthias and Savva, Manolis and Song, Shuran and Zeng, Andy and Zhang, Yinda},
  journal={arXiv preprint arXiv:1709.06158},
  year={2017}
}

@inproceedings{azuma2022scanqa,
  title={Scanqa: 3d question answering for spatial scene understanding},
  author={Azuma, Daichi and Miyanishi, Taiki and Kurita, Shuhei and Kawanabe, Motoaki},
  booktitle={proceedings of the IEEE/CVF conference on computer vision and pattern recognition},
  pages={19129--19139},
  year={2022}
}

@inproceedings{achlioptas2020referit3d,
  title={Referit3d: Neural listeners for fine-grained 3d object identification in real-world scenes},
  author={Achlioptas, Panos and Abdelreheem, Ahmed and Xia, Fei and Elhoseiny, Mohamed and Guibas, Leonidas},
  booktitle={European conference on computer vision},
  pages={422--440},
  year={2020},
  organization={Springer}
}

@article{fu2024scene,
  title={Scene-llm: Extending language model for 3d visual understanding and reasoning},
  author={Fu, Rao and Liu, Jingyu and Chen, Xilun and Nie, Yixin and Xiong, Wenhan},
  journal={arXiv preprint arXiv:2403.11401},
  year={2024}
}

@inproceedings{huang2024embodied,
  title={An embodied generalist agent in 3D world},
  author={Huang, Jiangyong and Yong, Silong and Ma, Xiaojian and Linghu, Xiongkun and Li, Puhao and Wang, Yan and Li, Qing and Zhu, Song-Chun and Jia, Baoxiong and Huang, Siyuan},
  booktitle={Proceedings of the 41st International Conference on Machine Learning},
  pages={20413--20451},
  year={2024}
}

@inproceedings{guo2023viewrefer,
  title={Viewrefer: Grasp the multi-view knowledge for 3d visual grounding},
  author={Guo, Zoey and Tang, Yiwen and Zhang, Ray and Wang, Dong and Wang, Zhigang and Zhao, Bin and Li, Xuelong},
  booktitle={Proceedings of the IEEE/CVF International Conference on Computer Vision},
  pages={15372--15383},
  year={2023}
}

@article{gholami2025spatial,
  title={Spatial reasoning with vision-language models in ego-centric multi-view scenes},
  author={Gholami, Mohsen and Rezaei, Ahmad and Weimin, Zhou and Mao, Sitong and Zhou, Shunbo and Zhang, Yong and Akbari, Mohammad},
  journal={arXiv preprint arXiv:2509.06266},
  year={2025}
}

@inproceedings{yin2025spatial,
  title={Spatial mental modeling from limited views},
  author={Yin, Baiqiao and Wang, Qineng and Zhang, Pingyue and Zhang, Jianshu and Wang, Kangrui and Wang, Zihan and Zhang, Jieyu and Chandrasegaran, Keshigeyan and Liu, Han and Krishna, Ranjay and others},
  booktitle={Structural Priors for Vision Workshop at ICCV'25},
  year={2025}
}

@article{gao2024embodiedcity,
  title={Embodiedcity: A benchmark platform for embodied agent in real-world city environment},
  author={Gao, Chen and Zhao, Baining and Zhang, Weichen and Mao, Jinzhu and Zhang, Jun and Zheng, Zhiheng and Man, Fanhang and Fang, Jianjie and Zhou, Zile and Cui, Jinqiang and others},
  journal={arXiv preprint arXiv:2410.09604},
  year={2024}
}

@inproceedings{li2023matrixcity,
  title={Matrixcity: A large-scale city dataset for city-scale neural rendering and beyond},
  author={Li, Yixuan and Jiang, Lihan and Xu, Linning and Xiangli, Yuanbo and Wang, Zhenzhi and Lin, Dahua and Dai, Bo},
  booktitle={Proceedings of the IEEE/CVF International Conference on Computer Vision},
  pages={3205--3215},
  year={2023}
}

@inproceedings{chu2024towards,
  title={Towards natural language-guided drones: GeoText-1652 benchmark with spatial relation matching},
  author={Chu, Meng and Zheng, Zhedong and Ji, Wei and Wang, Tingyu and Chua, Tat-Seng},
  booktitle={European Conference on Computer Vision},
  pages={213--231},
  year={2024},
  organization={Springer}
}

@inproceedings{caesar2020nuscenes,
  title={nuscenes: A multimodal dataset for autonomous driving},
  author={Caesar, Holger and Bankiti, Varun and Lang, Alex H and Vora, Sourabh and Liong, Venice Erin and Xu, Qiang and Krishnan, Anush and Pan, Yu and Baldan, Giancarlo and Beijbom, Oscar},
  booktitle={Proceedings of the IEEE/CVF conference on computer vision and pattern recognition},
  pages={11621--11631},
  year={2020}
}

@article{cai2025has,
  title={Has gpt-5 achieved spatial intelligence? an empirical study},
  author={Cai, Zhongang and Wang, Yubo and Sun, Qingping and Wang, Ruisi and Gu, Chenyang and Yin, Wanqi and Lin, Zhiqian and Yang, Zhitao and Wei, Chen and Shi, Xuanke and others},
  journal={arXiv preprint arXiv:2508.13142},
  volume={3},
  year={2025}
}

@article{yeh2025seeing,
  title={Seeing from another perspective: Evaluating multi-view understanding in mllms},
  author={Yeh, Chun-Hsiao and Wang, Chenyu and Tong, Shengbang and Cheng, Ta-Ying and Wang, Ruoyu and Chu, Tianzhe and Zhai, Yuexiang and Chen, Yubei and Gao, Shenghua and Ma, Yi},
  journal={arXiv preprint arXiv:2504.15280},
  year={2025}
}

@article{yang2025mmsi,
  title={MMSI-Bench: A Benchmark for Multi-Image Spatial Intelligence},
  author={Yang, Sihan and Xu, Runsen and Xie, Yiman and Yang, Sizhe and Li, Mo and Lin, Jingli and Zhu, Chenming and Chen, Xiaochen and Duan, Haodong and Yue, Xiangyu and others},
  journal={arXiv preprint arXiv:2505.23764},
  year={2025}
}

@article{liu2025spatial,
  title={Spatial-SSRL: Enhancing Spatial Understanding via Self-Supervised Reinforcement Learning},
  author={Liu, Yuhong and Zhang, Beichen and Zang, Yuhang and Cao, Yuhang and Xing, Long and Dong, Xiaoyi and Duan, Haodong and Lin, Dahua and Wang, Jiaqi},
  journal={arXiv preprint arXiv:2510.27606},
  year={2025}
}

@misc{openai2025gpt5systemcard,
  author       = {OpenAI},
  title        = {GPT-5 System Card},
  year         = {2025},
  month        = aug,
  url          = {https://openai.com/zh-Hans-CN/index/gpt-5-system-card/},
  note         = {Accessed: 2026-01-03}
}

@article{comanici2025gemini,
  title={Gemini 2.5: Pushing the frontier with advanced reasoning, multimodality, long context, and next generation agentic capabilities},
  author={Comanici, Gheorghe and Bieber, Eric and Schaekermann, Mike and Pasupat, Ice and Sachdeva, Noveen and Dhillon, Inderjit and Blistein, Marcel and Ram, Ori and Zhang, Dan and Rosen, Evan and others},
  journal={arXiv preprint arXiv:2507.06261},
  year={2025}
}

@misc{bai2025qwen3vltechnicalreport,
      title={Qwen3-VL Technical Report}, 
      author={Shuai Bai and Yuxuan Cai and Ke Zhu},
      year={2025},
      eprint={2511.21631},
      archivePrefix={arXiv},
      primaryClass={cs.CV},
      url={https://arxiv.org/abs/2511.21631}, 
}

@misc{wei2023skywork,
      title={Skywork: A More Open Bilingual Foundation Model}, 
      author={Tianwen Wei and Liang Zhao and Yahui Zhou},
      year={2023},
      eprint={2310.19341},
      archivePrefix={arXiv},
      primaryClass={cs.CL}
}

@misc{guo2025seed15vltechnicalreport,
      title={Seed1.5-VL Technical Report}, 
      author={Dong Guo and Faming Wu and Zuquan Song},
      year={2025},
      eprint={2505.07062},
      archivePrefix={arXiv},
      primaryClass={cs.CV},
      url={https://arxiv.org/abs/2505.07062}, 
}

@article{sensenova-si,
  title = {Scaling Spatial Intelligence with Multimodal Foundation Models},
  author = {Cai, Zhongang and Wang, Ruisi and Yang, Lei},
  journal = {arXiv preprint arXiv:2511.13719},
  year = {2025}
}
